\documentclass[a4paper,fleqn]{cas-dc}

\usepackage{hyperref}
\usepackage{xcolor}
\usepackage{multirow}
\usepackage{gensymb}
\usepackage{algorithm}
\usepackage{algpseudocode}
\usepackage{graphicx}
\usepackage{subcaption}

\usepackage[authoryear]{natbib}
\def\tsc#1{\csdef{#1}{\textsc{\lowercase{#1}}\xspace}}
\tsc{WGM}
\tsc{QE}
\begin{document}
\let\WriteBookmarks\relax
\def\floatpagepagefraction{1}
\def\textpagefraction{.001}

\shorttitle{}    

\shortauthors{}  

\title [mode = title]{Geographically-Weighted Weakly Supervised Bayesian High-Resolution Transformer for 200m Resolution Pan-Arctic Sea Ice Concentration Mapping and Uncertainty Estimation using Sentinel-1, RCM, and AMSR2 Data}  




\author[1]{Mabel Heffring}[
orcid=0009-0007-5069-7431,
]


\ead{mabel.heffring1@ucalgary.ca}


\credit{This work was supported by the Natural
Sciences and Engineering Research Council of Canada (NSERC) under Grant RGPIN-2019-06744.}

\affiliation[1]{organization={Department of Geomatics Engineering, Schulich School of Engineering, University of Calgary},
            addressline={2500 University Dr NW}, 
            city={Calgary},
%          citysep={}, % Uncomment if no comma needed between city and postcode
            postcode={T2N 1N4}, 
            state={AB},
            country={Canada}}

\author[1]{Lincoln Linlin Xu}[
orcid=0000-0002-3488-5199,
]

% Footnote of the second author
\fnmark[1]

\cormark[1]

% Email id of the second author
\ead{lincoln.xu@ucalgary.ca}



%\affiliation[2]{organization={},
%          citysep={}, % Uncomment if no comma needed between city and postcode

\cortext[1]{Corresponding author}

\fntext[1]{This work was supported by the Natural
Sciences and Engineering Research Council of Canada (NSERC) under Grant RGPIN-2019-06744.}


\begin{abstract}
Although high-resolution mapping of pan-Arctic sea ice with reliable corresponding uncertainty is essential for operational sea ice concentration (SIC) charting, it is a difficult task due to key challenges, such as the subtle nature of ice signature features, inexact SIC labels, model uncertainty, and data heterogeneity. This study presents a novel Bayesian High-Resolution Transformer approach for 200 meter resolution pan-Arctic SIC mapping and uncertainty quantification using Sentinel-1, RADARSAT Constellation Mission (RCM), and Advanced Microwave Scanning Radiometer 2 (AMSR2) data. First, to improve small and subtle sea ice feature (e.g., cracks/leads, ponds, and ice floes) extraction, we design a novel high-resolution Transformer model with both global and local modules that can better discern the subtle differences in sea ice patterns. Second, to address low-resolution and inexact SIC labels, we design a geographically-weighted weakly supervised loss function to supervise the model at region level instead of pixel level, and to prioritize pure open water and ice pack signatures while mitigating the impact of ambiguity in the marginal ice zone (MIZ). Third, to improve uncertainty quantification, we design a Bayesian extension of the proposed Transformer model, treating its parameters as random variables to more effectively capture uncertainties. Fourth, to address data heterogeneity, we fuse three different data types (Sentinel-1, RCM, and AMSR2) at decision-level to improve both SIC mapping and uncertainty quantification. The proposed approach is evaluated under pan-Arctic minimum-extent conditions in 2021 and 2025. Results demonstrate that the proposed model generates high-resolution SIC maps with up to 200 meter spatial resolution, achieving 0.70 overall feature detection accuracy using Sentinel-1 data, while also preserving pan-Arctic SIC patterns (Sentinel-1 R\textsuperscript{2} = 0.90 relative to the ARTIST Sea Ice product). Compared with alternative uncertainty quantification approaches, the Bayesian Transformer exhibits superior calibration and robust uncertainty estimates, with the lowest standard deviations ranging from 0.01\% to 0.33\% using Sentinel-1 data. Decision-level fusion integrates both SIC accuracy and uncertainty, prioritizing Sentinel-1, followed by RCM and AMSR2.
\end{abstract}

%\nocite{*}

% Keywords
\begin{keywords}
Pan-Arctic Sea Ice Concentration (SIC) \sep Uncertainty Quantification \sep Weak Supervision \sep Synthetic Aperture Radar (SAR) \sep Passive Microwave (PM) \sep Data Fusion
\end{keywords}

\maketitle

% Main text
\section{Introduction}
\label{introduction}

Accurate mapping of Pan-Arctic sea ice concentration (SIC) and evaluation of the SIC map uncertainties are essential for climate change studies, Arctic sea route navigation, and climate adaptation of the people and animals that call the Arctic home. Existing operational pan-Arctic SIC maps rely on manually-delineated regional ice charts and low-resolution satellite products. While sufficient for large-scale climate modeling, these methods lack the spatial detail required for safe navigation in the increasingly dynamic Arctic environment. Nevertheless, achieving high-resolution SIC maps with robust uncertainty quantification is a difficult task due to several key factors, i.e., the subtle nature of ice signature features, inexact SIC labels, model ambiguity, and data heterogeneity. 

First, high-resolution SIC products require the models to discern subtle differences in SIC classes and capture small ice features (e.g., ice floes, marginal ice zone (MIZ) boundaries, and cracks/leads on the ice pack). However, these are challenging tasks due to the complex and ever-changing marine environment and the noise effect on satellite sensors, i.e., Synthetic Aperture Radar (SAR) from RADARSAT Constellation Mission (RCM) and Sentinel-1, and Passive Microwave (PM) radiometry from the Advanced Microwave Scanning Radiometer 2 (AMSR2) \cite{ wang2016sea, malmgren2021convolutional}. Therefore, the development of advanced high-resolution deep learning approaches to extract discriminative and subtle sea ice features from SAR and AMSR2 images is a critical task. Over the past decade, numerous deep learning (DL) approaches have been proposed. Convolutional Neural Networks (CNN) \cite{wang2016sea, wulf2024panarctic, malmgren2021convolutional} are commonly applied to SIC mapping due to their efficiency, but their strong inductive bias and focus on local features make it challenging to capture global context in SAR and PM sea ice imagery. Moreover, CNN models tend to smooth out small spatial features due to their weight-sharing mechanism. In contrast, Transformers are more flexible and globally aware \cite{jiang2024SICFormer,He2025physically, chen2025sea, ren2025sicnet, he2026explainable}, making them a viable choice for learning the subtle nature of ice signature features on a pan-Arctic scale. To address the essential need for improved sea ice feature extraction, we therefore propose a novel high-resolution Transformer architecture.

Second, training high-resolution models using low-resolution ground truth is a challenging research issue. As it is difficult to conduct field studies and obtain large amounts of ground samples for training DL models, one feasible approach is to use low-resolution SIC product to train high-resolution DL models. However, the inexactness and inaccuracy of existing low-resolution SIC products pose significant challenges for effectively supervising high-resolution SIC models. Gridded SIC products derived from AMSR2, using retrieval algorithms such as NASA Team (NT) \cite{NTdetails} and ARTIST Sea Ice (ASI) \cite{spreen2008asi}, are typically produced at coarse spatial resolutions with order of several kilometers. Furthermore, the accuracy of gridded SIC labels depends on their geographic location. More homogeneous surface types, such as open water and consolidated pack ice, are generally well represented at coarse spatial resolutions. However, performance degrades within the MIZ, where strong spatial heterogeneity and sharp concentration gradients occur over short distances. Various weakly supervised learning strategies have been proposed to address inexact SIC labels \cite{cooke2019estimating, gon2021fine, chen2024weakly}, yet geographically-weighted weakly supervised SIC mapping approaches that account for not only the inexactness of SIC products but also their location-dependent nature at pan-Arctic scale are still insufficiently researched.  

Third, high-resolution pan-Arctic SIC mapping requires reliable uncertainty estimation to quantify the model's confidence level and data ambiguities.  Variational inference methods such as Monte Carlo (MC) simulation, ensemble generation, and Bayes by Backpropagation (BBB) \cite{ABDAR2021243} are commonly used for DL uncertainty quantification in Earth observation. Although MC simulation and ensemble generation can provide meaningful uncertainty estimates, these values are heuristic approximations that may not represent the true posterior distribution. Bayesian DL architectures that incorporate BBB explicitly estimate the model parameters as random variables \cite{aires2004neural,asadi2020evaluation}, resulting in more reliable and better calibrated uncertainties. Therefore, incorporating Bayesian uncertainty estimation into high-resolution DL models at pan-Arctic scale is an important research issue. 

Fourth, SIC mapping is complicated by Sentinel-1, RCM, and AMSR2 data heterogeneity. Although SAR systems enable high-resolution ice mapping due to their fine spatial resolution, they do not provide complete daily pan-Arctic coverage and have strong noise patterns, particularly in open water areas (e.g., incidence angle effect and thermal noise). PM systems, despite having low spatial resolution, provide complete daily pan-Arctic coverage with clear delineation between ice and open water. Various data fusion methods have been developed to combine the strengths of SAR and PM systems for sea ice mapping, e.g., \cite{wang2016improved, wulf2024panarctic}. These methods are feature-level fusion approaches. The decision-level fusion, where fusion is performed after classification for each respective system, has not been fully explored to improve high-resolution pan-Arctic sea ice mapping.

This study therefore presents a novel Bayesian Transformer approach for pan-Arctic SIC mapping and uncertainty quantification using Sentinel-1, RADARSAT Constellation Mission (RCM), and Advanced Microwave Scanning Radiometer 2 (AMSR2) data with the following contributions:

\begin{itemize}
    \item First, to improve small ice feature extraction and enhance spatial resolution, we design a novel high-resolution Transformer model with both global and local modules that can better discern the subtle differences in sea ice patterns and capture small and weak ice features.
    \item Second, to address inexact SIC labels whose accuracy and trustworthiness are location-dependent, we design a geographically-weighted weakly supervised loss. This approach supervises the proposed high-resolution Transformer model at region level instead of pixel level and prioritizes pure open water and ice pack signatures while mitigating the impact of ambiguity in the MIZ.
    \item Third, to improve uncertainty quantification, we design a Bayesian extension of the proposed Transformer model, treating its parameters as random, normally distributed variables to not only more effectively capture uncertainties, but also improve the accuracy of the high-resolution Transformer model. 
    \item Fourth, to address data heterogeneity, we fuse three different data types (Sentinel-1, RCM, and AMSR2) at decision-level to improve both pan-Arctic SIC mapping and uncertainty quantification, leading to high-quality SIC map at pan-Arctic scale.
\end{itemize}

 This paper proceeds as follows. Section \ref{related_works} summarizes related works corresponding to this study. Section \ref{data} describes the data and study area. Section \ref{methodology} outlines the details of the Bayesian High-Resolution Transformer architecture and implementation. Section \ref{experiments} presents the results and analysis. Section \ref{conclusion} concludes this study.

 \section{Related Works}
\label{related_works}

\subsection{Evolution of Deep Learning-based SIC Mapping}

Arctic SIC mapping is commonly posed as a regression task \cite{Li2024advancing}, where the fraction of ice within a pixel ranges from 0\% to 100\%. Sentinel-1, RCM, and AMSR2 provide all-weather, day-and-night observations that are well suited for SIC mapping, yet it is challenging to discern subtle differences in sea ice signatures across the diverse pan-Arctic region and capture small ice features (e.g., ice floes, marginal ice MIZ boundaries, and cracks/leads on the ice pack). SAR backscatter from sea ice or water is influenced by the radar frequency and incidence angle, environmental conditions like wind speed, and characteristics of the scatterer \cite{ochilov2012operational,wang2016sea}. Even discrimination of open water (0\% SIC) and ice covered areas from SAR imagery is challenging in certain conditions, where the incidence angle and wind can cause water to appear as bright as sea ice \cite{wang2016sea, malmgren2021convolutional}. AMSR2 brightness temperatures are better at discerning between open water and ice  \cite{malmgren2021convolutional}, however, environmental factors like cloud and fog can reduce visibility and degrade the quality of higher frequency channels. Consequently, many studies aim to develop robust methods for detecting subtle variations in ice signature features to improve SIC mapping.

Historically, sea ice classification was performed using texture features, such as Markov Random Fields (MRF), Gabor features, and gray level co-occurrence matrices (GLCM) \cite{ochilov2012operational, clausi2001comparison, clausi2004comparing, liu2014svm, wenbo2015sea}. These features classify different textures in SAR imagery, which is useful for discerning between smooth and rough ice. Classical machine learning studies have leveraged these features for sea ice classification using supervised Support Vector Machine (SVM) \cite{liu2014svm, jiang2022sea} and Random Forest (RF) models \cite{jiang2022sea}, as well as an unsupervised MRF-based segmentation model \cite{huawu2005unsupervised}. As these texture-based approaches rely on time-consuming manual feature delineation \cite{degelis2021prediction}, they are less suitable for operational SIC mapping.

In recent years various deep learning (DL) approaches have been proposed for Arctic SIC mapping. Many Convolutional Neural Networks (CNN) have been applied or adapted, including classic ConvNet \cite{wang2016sea,wang2017sea, wulf2024panarctic}, U-Net \cite{andersson_seasonal_2021, ren2022datadriven, stokholm2022ai4seaice, wang_arctic_2021}, ASPP-CNN \cite{malmgren2021convolutional,tamber2022accounting}, AlexNet \cite{chen2023uncertainty}, DenseNet \cite{cooke2019estimating}, Fully Convolutional Networks (FCN) \cite{degelis2021prediction}, and DeepLabv3 \cite{jalayer_enhancing_2025}. Shared weights, limited receptive fields, and pooling layers make CNN-based methods efficient, compact, and focused on local feature extraction. However, these model design elements introduce strong inductive bias that ignore global, long-range dependencies. This makes it challenging to extract local sea ice features like leads and floes while accurately modeling global, pan-Arctic SIC patterns using CNN-based approaches.

Transformer-based approaches are more flexible and globally aware \cite{jiang2024SICFormer,He2025physically, chen2025sea, ren2025sicnet, he2026explainable}, yet they have not been fully explored for high-resolution, pan-Arctic SIC mapping. The finest known scale achieved by a Transformer-based SIC mapping model is proposed in \cite{He2025physically} with a resolution of 0.1$\degree$ $\times$ 0.1$\degree$. This resolution would not allow for the detection of small sea ice features, i.e., leads and floes, that can span narrow widths of meters to tens of meters. Therefore, more research is required to assess Transformer models for high-resolution SIC mapping.

\subsection{Deep Learning Training Strategies for SIC Mapping}

Manually-delineated ice charts are commonly used as training labels for SIC mapping in both classification \cite{malmgren2021convolutional, wulf2024panarctic} and regression \cite{wang2016sea, wang2017sea} contexts. While ice charts are readily available and can infuse large-scale expert knowledge into DL training strategies, these region-level labels do not represent small-scale sea ice features. Various gridded SIC products derived from AMSR2 have also been used as training labels, including NASA Team \cite{rad2021sea}, ASI \cite{spreen2008asi,cooke2019estimating} and National Snow and Ice Data Center (NSIDC) \cite{chen2025sea} SIC products. These products offer greater detail than ice charts, yet they remain coarse in spatial resolution and are particularly inexact within the MIZ, where spatial heterogeneity is high. To achieve higher resolution SIC maps, researchers have implemented weak supervision where consistency with ground truth labels is maintained at the region level instead of the pixel level \cite{cooke2019estimating, gon2021fine, chen2024weakly}. Training methods for improved MIZ mapping have also been proposed, including curriculum learning \cite{rad2021sea} and rescaling of the ice chart using SAR imagery \cite{tamber2022accounting}. However, these approaches fail to account for the geographical dependency of inexact SIC labels.

Geographically-weighted regression \cite{charlton1998geographically}, which explicitly accounts for spatial non-stationarity by allowing regression coefficients to vary across space, has been combined with DL to improve land surface temperature \cite{jia2021predicting} and air quality \cite{li2020geographically} mapping. Geographically-weighted training strategies have also been applied by integrating spatial context into the loss function. \cite{wang2021neighborloss} dynamically weighted pixels according to the consistency of their immediate neighbours, while \cite{wang2024spatial} improved geological mapping by integrating domain-specific constraints, such as depth and structural proximity, to ensure predictions align with true physical properties. Therefore, a geographically-weighted and weakly supervised learning strategy can improve SIC mapping by addressing the inherent spatial variability of ice and water signatures.

\subsection{Uncertainty Quantification in Deep Learning}

Quantifying uncertainty in DL-derived SIC maps is essential for understanding potential inaccuracies, biases, and errors in the estimates, while also revealing model confidence and decision-making behavior. Moreover, uncertainty quantification is required for effective integration of SIC products into ice forecasting systems via operational data assimilation and decision-making tools \cite{wulf2024panarctic}. Variational inference (VI) is often used for uncertainty estimation in large-scale models with non-linear applications because sampling the posterior distribution is simple and efficient \cite{ABDAR2021243}. VI is applied for uncertainty estimation by outputting multiple predictions for the same input dataset and quantifying the dispersion of these predictions. VI methods used in remote sensing applications include Monte Carlo (MC) simulation \cite{PADARIAN2022assessing,MARTINEZFERRER2022quantifying,wernecke2024estimating}, ensemble generation \cite{Ferguson2010quantifying,sahay2021hyperspectral, pires2023modelensemble,DUGASSA2026deep}, and Bayes by backpropagation (BBB) \cite{aires2004neural,asadi2020evaluation,chen2023uncertainty, chen2023predicting}.

MC simulation introduces variability into a deterministic system using random sampling. In DL models, MC dropout removes random nodes or connections in the neural network during inference \cite{PADARIAN2022assessing,MARTINEZFERRER2022quantifying,wernecke2024estimating}. Although MC dropout can be more efficient than traditional Bayesian inference, it is prone to poor calibration \cite{MARTINEZFERRER2022quantifying,wang_uncertainty_2025}. For ensemble generation, multiple models are trained using different priors or training data, and the output predictions are compared \cite{Ferguson2010quantifying,sahay2021hyperspectral, pires2023modelensemble,DUGASSA2026deep}. This method lacks formal probabilistic interpretability \cite{wang_uncertainty_2025}, yet it has achieved higher accuracies in Sentinel-2 super resolution than deterministic methods \cite{iaguru2023uncertainty}. BBB is commonly applied in Bayesian Neural Networks (BNN), where the network weights are assumed to be probability distributions instead of single values and the weight distributions are sampled during inference \cite{aires2004neural,asadi2020evaluation,chen2023uncertainty, chen2023predicting}. Incorporating Bayesian uncertainty estimation into Multilayer Perceptron (MLP) models has improved model performance and reduced misclassification for detection of ice and water in SAR imagery \cite{asadi2020evaluation}. However, BBB has never been incorporated into Transformer-based SIC mapping methods. BBB achieved more reliable uncertainty estimates with lower calibration error when compared to MC dropout and ensemble generation for lithological mapping \cite{wang_uncertainty_2025}. A comparison of these DL uncertainty quantification methods has not yet been performed for pan-Arctic SIC mapping.

\subsection{Data Fusion for Deep Learning-based SIC Mapping}

Data fusion methodologies for SIC mapping can be structured into two primary levels: feature-level and decision-level \cite{schmitt2016datafusion}. Feature-level fusion combines latent features from each respective sensor system before classification, while decision-level fusion combines the final classification outputs. Recent DL-based feature-level fusion methods for SIC mapping involve the concatenation of SAR bands and PM brightness temperatures as input to the model \cite{wulf2024panarctic} or tailor-designed multi-branch architecture \cite{malmgren2021convolutional}. These methods exploit the correlation across different data types to learn subtle sea ice signature features. However, feature-level fusion can be heavily influenced by noise characteristics and inconsistencies across different sensors. Vastly different resolutions in SAR and PM data can degrade fine scale SIC predictions when using feature-level fusion.

Probabilistic decision-level methods like Bayesian \cite{wang2016improved} and Graph Laplacian \cite{khachatrian2023sar} fusion schemes have been proposed for sea ice classification from SAR and PM imagery, yet these methods require manual delineation of ice features. DL-based decision-level fusion methods can use fuzzy rule-based logic to aggregate classification outputs according to the capabilities of each classifier \cite{fauvel2006decision,zeng2006comparison}, leveraging the unique strengths of each data source. As decision-level fusion has not been explored for SIC mapping, it offers a unique opportunity for both daily full coverage and high-resolution maps.

\section{Data}
\label{data}

\subsection{Active and Passive Satellite Imagery}

\begin{table}[]
\renewcommand{\arraystretch}{1.2}
\centering
\caption{Overview of Sentinel-1, RCM, and AMSR2 Data Specifications.}
\begin{tabular}{l|l|l|l}
\toprule
Sensor                                                        & Sentinel-1                                                & RCM                                                                                              & AMSR2                                                                                        \\ \hline
\begin{tabular}[c]{@{}l@{}}Product\\ Level\end{tabular}       & \begin{tabular}[c]{@{}l@{}}Level-1\\GRD\end{tabular}                                               & \begin{tabular}[c]{@{}l@{}}Level-1\\GRD\end{tabular}                                                                                      & Level-1B                                                                                     \\ \hline
\begin{tabular}[c]{@{}l@{}}Acquisition\\ Mode(s)\end{tabular} & \begin{tabular}[c]{@{}l@{}}Extra Wide\\ (EW)\end{tabular} & \begin{tabular}[c]{@{}l@{}}Low \\ Resolution, \\ Medium \\ Resolution, \\ Low Noise\end{tabular} & \begin{tabular}[c]{@{}l@{}}Conical \\ scanning\end{tabular}                                  \\ \hline
\begin{tabular}[c]{@{}l@{}}Frequency\\ Band(s)\end{tabular}   & C-Band                                                    & C-Band                                                                                           & \begin{tabular}[c]{@{}l@{}}18.7, 23.8, \\ 36.5, \\ 89.0 GHz\end{tabular}                     \\ \hline
Polarizations                                                 & HH, HV                                                    & HH, HV                                                                                           & H, V                                                                                         \\ \hline
\begin{tabular}[c]{@{}l@{}}Ground\\ Resolution\end{tabular}   & 20 x 40 m                                                 & \begin{tabular}[c]{@{}l@{}}50 m -\\  100 m\end{tabular}                                                                                     & \begin{tabular}[c]{@{}l@{}}14 x 22 km, \\ 15 x 26 km, \\ 7 x 12 km, \\ 3 x 5 km\end{tabular}\\ \hline
\begin{tabular}[c]{@{}l@{}}Daily\\ Raw Data\end{tabular}   & $\sim$ 30 GB                                                 & \begin{tabular}[c]{@{}l@{}}$\sim$ 200 GB\end{tabular}                                                                                     & \begin{tabular}[c]{@{}l@{}}$\sim$ 1 GB\end{tabular}\\
\bottomrule
\end{tabular}
\label{tab:data_spec}
%\vspace{-4mm}
\end{table}

The satellite data for this study includes SAR from Sentinel-1 and RCM and PM from AMSR2. Table \ref{tab:data_spec} summarizes the product level, acquisition mode, frequencies, polarizations, ground resolution, and daily raw data sizes for the prepared Sentinel-1, RCM, and AMSR2 satellite imagery. The general preprocessing pipeline for preparing analysis-ready SAR imagery includes (1) downloading data via API services, (2) radiometric calibration to extract backscatter coefficient, $\sigma^0$, from raw digital numbers, (3) multi-looking to 200 m resolution to reduce speckle noise, (4) geocoding, and (5) combining all scenes into a pan-Arctic mosaic. For Sentinel-1, an additional pre-processing step is included to remove the incidence angle effect and thermal noise. Sentinel-1 is collected in Extra Wide (EW) mode with HH and HV polarization and 20 m x 40 m ground resolution. RCM is collected in low resolution (100 m resolution), medium resolution (50 m), and low noise (100 m) modes with HH and HV polarization. An additional cross-polarization layer is derived for both SAR systems to enhance physical structures and reduce the impact of noise \cite{komarov2022ocean}. 

The preprocessing pipeline for AMSR2 imagery includes (1) downloading AMSR2 Level-1B swath data via API services, (2) geocoding, (3) combining all scenes into a pan-Arctic mosaic, and (4)  resampling to match the 200 m resolution of the SAR data. AMSR2 Brightness Temperatures are collected for 18.7 (14 x 22 km resolution), 23.8 (15 km x 26 km), 36.5 (7 km x 12 km), and 89.0 GHz (3 x 5 km) frequencies with H and V polarization. 18.7, 23.8, and 36.5 GHz channels are used to derive SIC labels using the NASA Team algorithm \cite{NTdetails}. The 89.0 GHz channels are used as input for training.

While Sentinel-1 and RCM are high-resolution with partial daily spatial coverage, AMSR2 has lower resolution with complete coverage. AMSR2 also has lower cloud permeability at higher frequencies. Therefore, it is important to develop fusion techniques that leverage the strengths of these datasets for SIC mapping and uncertainty quantification.

\subsection{Low-Resolution SIC Products and Ice Charts}
The Bayesian high-resolution Transformer model is weakly-supervised using the NASA Team SIC product \cite{NTdetails} and NIC daily ice charts. The NASA Team algorithm determines SIC based on ratios of PM brightness temperatures using tie points for pure surface types, including open water and ice pack. While this approach is relatively insensitive to absolute brightness temperature variability, it is less effective at discriminating among ice types during the melt season and tends to underestimate thin ice. Furthermore, because the product is derived from the lower-frequency 18.7, 23.8, and 36.5 GHz channels, its spatial resolution is limited. The NIC daily ice charts are manually-delineated by ice experts who analyze satellite imagery from different sources, including both SAR and PM systems. These charts offer timely, daily assessments of sea ice conditions. However, they only provide region level detail and are subject to potential human error. Therefore, a geographically-weighted weakly supervised training strategy is necessary to account for inexact labels and provide accurate, high-resolution SIC estimates.

SIC accuracy is evaluated using pan-Arctic 3125 m SIC products derived from the ASI algorithm with AMSR2 data \cite{spreen2008asi}. This algorithm also utilizes the 89.0 GHz channels to achieve higher resolution with a weather filter to address atmospheric interference. According to \cite{spreen2008asi}, the systematic difference between NASA Team and ASI is approximately -2 $\pm$ 8.8\%. As the ASI product is also derived from AMSR2 and represents one of the highest-resolution operational SIC products currently available, it provides a suitable benchmark for validation in this study.

\section{Methodology}
\label{methodology}
\subsection{High-Resolution Transformer Architecture}

\begin{figure*}[!htbp]
    \centering
    \includegraphics[width=\linewidth]{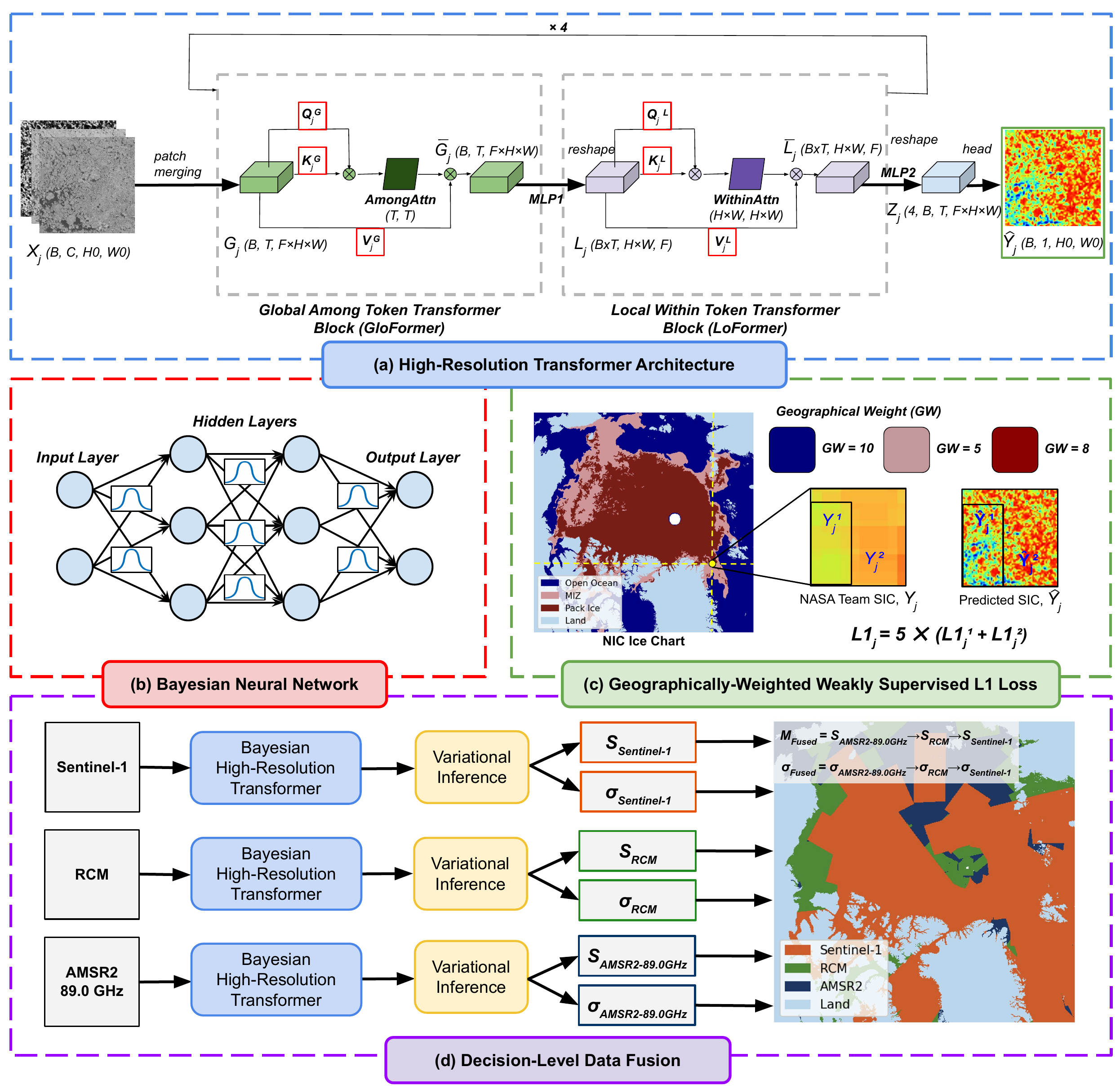}
    \vspace{-0.3cm}
    \caption{Overview of our key innovations for pan-Arctic SIC mapping with corresponding uncertainty quantification. The proposed model utilizes a \textbf{(a) High-Resolution Transformer} architecture, $f^{\omega}(.)$, with an among-token Transformer block for modeling global context (GloFormer) and a within-token Transformer block for modeling local detail (LoFormer). The high-resolution Transformer becomes a \textbf{(b) Bayesian Neural Network (BNN)} by assuming the model parameters in the attention mechanisms are probabilistic, as highlighted in red. The primary training strategy is a \textbf{(c) geographically-weighted weakly supervised $\mathcal{L}_{L1-GW}$ loss}, where the difference between the predicted SIC and NASA Team ground truth is evaluated at the cluster level rather than the high-resolution pixel level. Geographical weights dynamically scale the importance of each sample according to the NIC ice chart. Individual Bayesian Transformer models are trained using Sentinel-1, RCM, and AMSR2 89.0GHz in parallel. After variational inference, \textbf{(d) decision-level data fusion} is used to combine the SIC and corresponding uncertainty maps.}
    \label{fig:Transformer}
\end{figure*}

To capture both global sea ice context information and local small sea ice features, we design a novel high-resolution Transformer model with both a global among-token Transformer block (i.e., GloFormer) and a local within-token Transformer block (i.e., LoFormer), where the GloFormer focuses on learning large-scale correlation among tokens and LoFormer focuses on learning fine-grain spatial correlations among patches within a token. This novel high-resolution Transformer not only improves discernment of subtle SIC signatures, but also enhances the capturing of small and weak ice features, e.g., ice floes, cracks/leads and ice boundaries in the MIZ. 

The Bayesian High-Resolution Transformer is denoted as a regression function,
\begin{equation}
\ \ \ \ \ \ \ \ \ \ \ \ \ \ \ \ \ \ \ \ \ \ \ y = f^{\omega}(x)
\label{transformer}
\end{equation}
\noindent which derives SIC, $y$, from input data, $x$, by optimizing model parameters, $\omega$, through training.  

In Figure \ref{fig:Transformer}(a), the input data for the $j^{th}$ batch is $X_j \in \mathbb{R}^{B, C, H0, W0}$ where $B$ is the batch size, $C$ is the channel dimension, and $H0, W0$ are the height and width of the input chips, respectively. Patch merging is used to generate patch tokens ($G_j \in \mathbb{R}^{B, T, F \times H \times W}$) \cite{Liu_2021_ICCV}, where $T$ is the number of tokens, $F$ is the hidden dimension, and $H,W$ are the height and width of each token, respectively. Next, these patch tokens are passed sequentially through the GloFormer and LoFormer blocks to learn both global and local SIC patterns, respectively.

\subsubsection{Global Among-Token Transformer Block}
The among-token Transformer block (GloFormer) models the global context using a multi-head self-attention mechanism across all tokens. For the first head, the output is defined as:

\begin{equation}
\bar{G}_j^1 = \text{softmax}\Bigg(\frac{Q_j^G (K_j^G)^\top}{\sqrt{d_k^G}}\Bigg) V_j^G = (AmongAttn)V_j^G
\label{global_attn_eq}
\end{equation}

\noindent where $Q_j^G = G_j \mathcal{W}^{Q_j^G}, K_j^G = G_j \mathcal{W}^{K_j^G}, V_j^G = G_j \mathcal{W}^{V_j^G} \in \mathbb{R}^{B\times T\times d_k^G}$ are queries, keys, and values of the tokens, respectively, $d_k^G = (F\times H\times W) / h$ is the GloFormer key dimension, $h$ is the number of heads, and $AmongAttn$ are the attention scores. $\mathcal{W}^{Q_j^G}$, $\mathcal{W}^{K_j^G}$, and $\mathcal{W}^{V_j^G}$ are the projection weights of $Q_j^G$, $K_j^G$, and $V_j^G$, respectively. Concatenated and projected heads, $[\bar{G}_j^1, \bar{G}_j^2,...,\bar{G}_j^h]$, yield $\bar{G}_j \in \mathbb{R}^{B, T, F \times H \times W}$.

\subsubsection{Local Within-Token Transformer Block}

The within-token Transformer block (LoFormer) is designed to learn the local features using a multi-head self-attention mechanism across the patches inside each token. The output $\bar{G}_j$ is reshaped to $L_j \in \mathbb{R}^{B \times T, H \times W, F}$, and the LoFormer attention matrix for the first head is defined as:

\begin{equation}
\bar{L}_j^1 = \text{softmax}\Bigg(\frac{Q_j^L (K_j^L)^\top}{\sqrt{d_k^L}}\Bigg) V_j^L = (WithinAttn)V_j^L
\label{local_attn_eq}
\end{equation}

\noindent where $Q_j^L = L_j \mathcal{W}^{Q_j^L}, K_j^L = L_j \mathcal{W}^{K_j^L}, V_j^L = L_j \mathcal{W}^{V_j^L} \in \mathbb{R}^{B \times T \times d_k^L}$ are queries, keys, and values of the inner patches, respectively, $d_k^L = F / h$ is the LoFormer key dimension, $h$ is the number of heads, and $WithinAttn$ are the attention scores. $\mathcal{W}^{Q_j^L}$, $\mathcal{W}^{K_j^L}$, and $\mathcal{W}^{V_j^L}$ are the projection weights of $Q_j^L$, $K_j^L$, and $V_j^L$, respectively. Concatenated and projected heads, $[\bar{L}_j^1, \bar{L}_j^2,...,\bar{L}_j^h]$, yield $\bar{L}_j \in \mathbb{R}^{B \times T, H \times W, F}$.

A Multilayer Perceptron (MLP) after each Transformer block further improves complex pattern learning and memorization. The sequential GloFormer and LoFormer blocks are repeated four times to capture both high and low frequency features. To determine the SIC output after the final repetition, $\hat{Y}_j \in \mathbb{R}^{B, 1,H0,W0}$, the LoFormer attention matrix $\bar{L}_j$ is reshaped to $Z_j \in \mathbb{R}^{4, B, T, F \times H \times W}$ and an interpolation head is applied.

\subsection{Geographically-Weighted Weakly Supervised L1 Loss}
To address the inexactness of the NASA Team SIC product and its spatial-dependent trustworthiness, we design a novel  geographically-weighted weakly supervised $\mathcal{L}_{L1-GW}$ loss, as illustrated in Figure \ref{fig:Transformer}(c). 

To address the low-resolution and inexact NASA Team SIC product, the proposed model is supervised at region or cluster-level instead of pixel-level, which can better avoid the potential bias and errors of the SIC product at pixel level. More specifically, the difference between the predicted SIC and NASA Team ground truth is measured for each SIC cluster/region instead of each high-resolution pixel. This weak supervision method promotes general consistency with the low resolution SIC labels and reduces the impact of noisy labels, while also allowing the model to develop local fine-grained detail. 

SIC clusters are determined according to the NASA Team SIC, $Y^k$, for $k \in \{0, ..., K\}$, where $K$ is the total number of clusters. Cluster boundaries, $[k*10, (k+1)*10)$, ranging from 0\% to 100\% are used to filter the NASA Team SIC for each cluster. The corresponding high-resolution predicted SIC pixels, $\mathbf{\hat{Y}}^k \in \mathbb{R}^{N_k}$, are aggregated and averaged for each cluster as shown below, where $N_k$ is the total number of high-resolution predicted SIC pixels that belong to the current cluster. 

\begin{equation}
\mathcal{F}(\mathbf{\hat{Y}}^k) = \frac{1}{N_k} \sum_{i=1}^{N_k} \hat{y_i}
\end{equation}

The geographically-dependent weight of each sample, $GW$, is determined according to the daily U.S NIC ice charts, where more confident regions like open water and ice pack are weighted higher than the less confident MIZ. Therefore, the relative importance of a training sample is determined by its latitude and longitude. Geographically weighting the loss also accounts for the known limitations of the NASA Team SIC used for training, particularly underestimation of thin ice. The geographically-weighted weakly supervised $\mathcal{L}_{L1-GW}$ loss, for one sample is calculated as follows:

\begin{equation}
\mathcal{L}_{L1-GW}= \sum_{k=1}^{K} GW * \lVert \mathcal{F}(\mathbf{\hat{Y}}^k ) - Y^k \rVert_{1}
\label{L1_eq}
\end{equation}

For samples located in open water there is only one cluster corresponding to 0\% concentration. The total batch loss is calculated by aggregating $\mathcal{L}_{L1-GW}$ for all constituent samples. The L1 norm is chosen over the common L2 (MSE) function to avoid bias towards lower concentrations \cite{rad2021sea}. This loss is also used to determine the best epoch during validation. 

\subsection{Bayesian Transformer Model}

The high-resolution Transformer model becomes a Bayesian Neural Network (BNN) that can quantify epistemic uncertainty by assuming a probability distribution over the model parameters, $\omega$ \cite{ABDAR2021243}, as illustrated in Figure \ref{fig:Transformer}(b). For this study, a Gaussian likelihood for regression,  $p(y \mid x, \omega)$, is defined as follows:

\begin{equation}
    p(y \mid x, \omega) = \mathcal{N}\big(y \,;\, f^{\omega}(x), \tau^{-1} I\big)
    \label{gaussian_likelihood_eq}
\end{equation}
\noindent where $\tau$ is the model precision. By applying Bayes theorem, the posterior distribution for the training dataset \(\mathcal{D}=\{X,Y\}\) can be determined as follows:
\begin{equation}
    p(\omega \mid X,Y) = \frac{p(Y|X, \omega) \, p(\omega)}{p(Y|X)}
\end{equation}
The posterior distribution, $p(\omega \mid X,Y)$, is approximated by variational distribution, $q_{\theta}(\omega)$, as it cannot be derived explicitly. Therefore, the secondary learning objective of the Bayesian Transformer is to approximate a distribution that is close to the true posterior distribution found by the model through BBB, with respect to the variational parameters, $\theta$. Assuming a diagonal Gaussian variational posterior, $p(\omega) = \mathcal{N}(0, I)$, the Kullback-Leibler (KL) divergence loss is used to minimize the difference between the two distributions as follows: 
\begin{equation}
\mathcal{L}_{\text{KL}}
= \mathrm{KL}\!\big(q_{\theta}(\omega) \,\|\, p(\omega)\big)
= \int q_{\theta}(\omega) \, \log \frac{q_{\theta}(\omega)}{p(\omega)} \, d\omega
\end{equation}
The final loss for multitask learning is calculated as the sum of the geographically-weighted $\mathcal{L}_{L1-GW}$ loss and KL divergence loss, $\mathcal{L}=\mathcal{L}_{L1-GW} + \mathcal{L}_{KL}$. After training, the approximate variational distribution is randomly sampled to obtain the model parameters for a set number of inferences. The predictive SIC sample mean, $\mu(x)$, and variance (that is used to quantify uncertainties), $\sigma^{2}(x)$, are derived from these inferences as follows:
\begin{equation}
\mu(x) := \mathbb{E}_{q_\theta(\omega)} \big[ f^{\omega}(x) \big]
\label{eq_mean}
\end{equation}
\begin{equation}
\sigma^{2}(x) := \operatorname{Var}_{q_\theta(\omega)} \big[ f^{\omega}(x) \big]
\label{eq_variance}
\end{equation}

\subsection{Other Uncertainty Estimation Approaches}

\begin{figure}[]
    \begin{subfigure}[b]{\linewidth}
         \centering
         \includegraphics[width=0.75\linewidth]{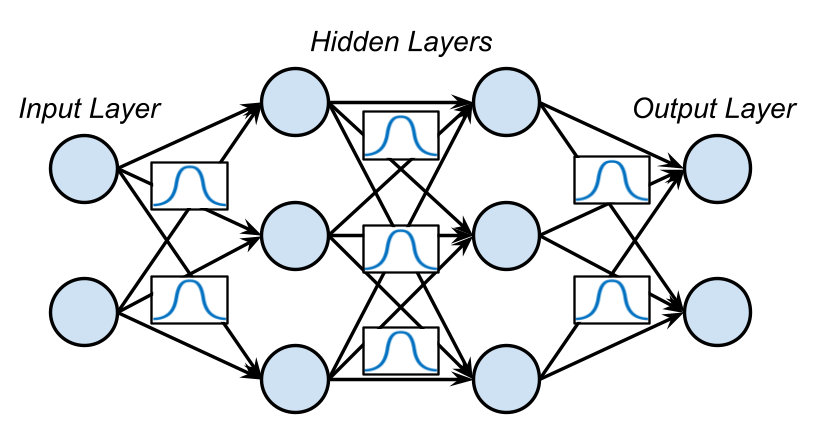}
         %\caption{\textbf{Bayesian Neural Network (BNN)} where the model parameters are assumed to be probabilistic.}
         \caption{Bayesian Neural Network (BNN)}
    \end{subfigure}

    \begin{subfigure}[b]{\linewidth}
         \centering
         \includegraphics[width=0.75\linewidth]{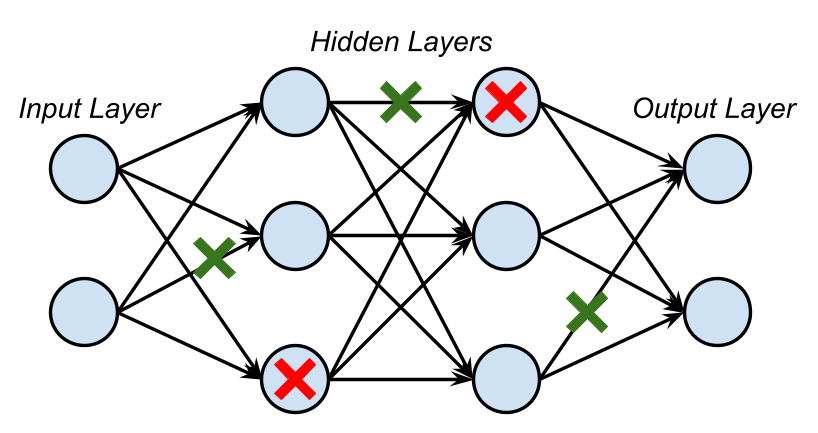}
         %\caption{\textbf{MC dropout} where red denotes masked nodes and green denotes masked connections. With each variational inference the dropout configuration changes, introducing randomness into the neural network.}
         \caption{Monte Carlo Dropout}
    \end{subfigure}
    
    \begin{subfigure}[b]{\linewidth}
         \centering
         \includegraphics[width=0.9\linewidth]{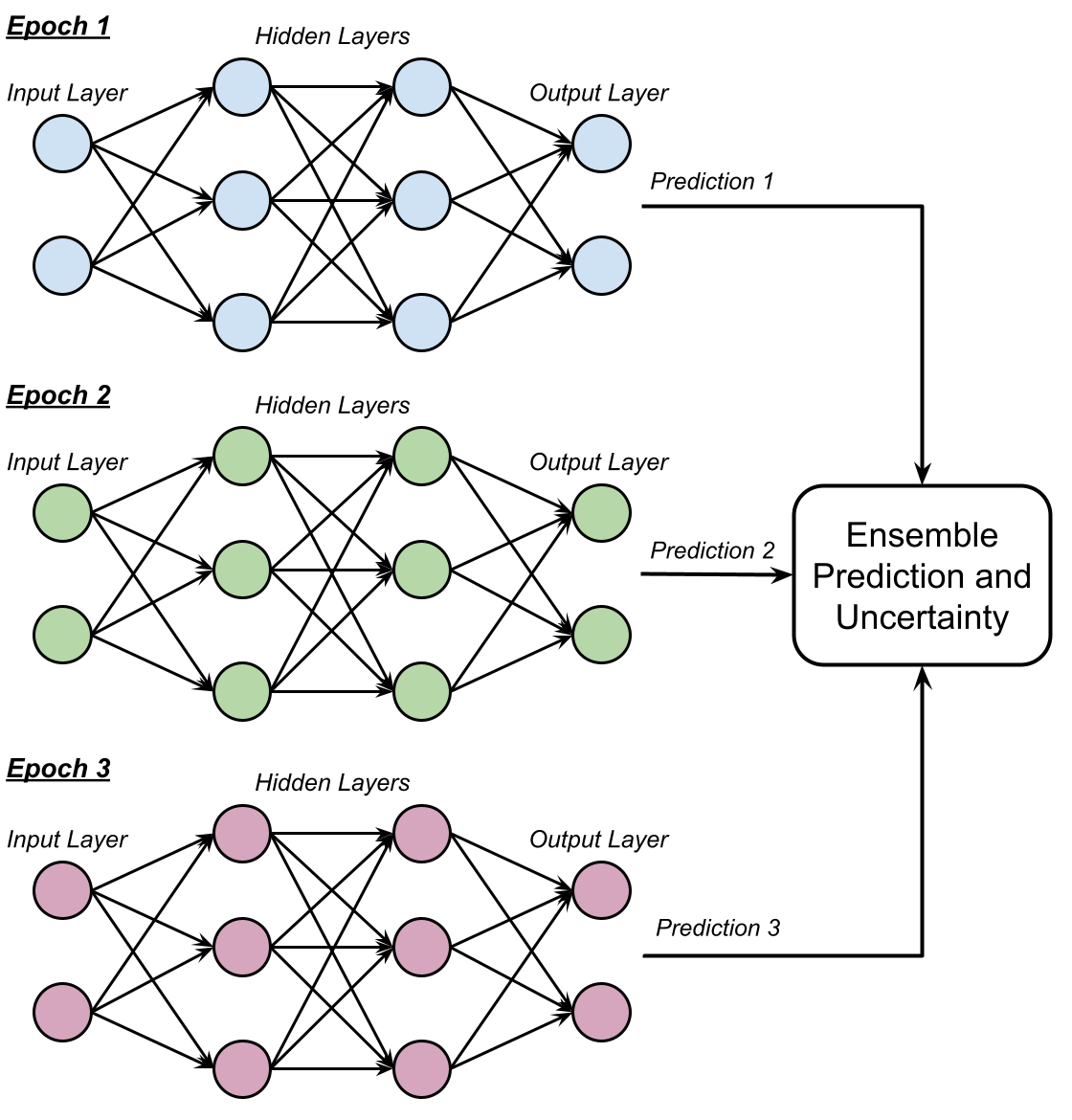}
         %\caption{\textbf{Epoch ensemble generation} where model parameters at each epoch are treated as independent ensemble members.}
         \caption{Epoch Ensemble Generation}
    \end{subfigure}

    \caption{Theoretical Comparison of Epistemic Uncertainty Quantification Approaches. Figure (a) illustrates the \textbf{Bayesian Neural Network (BNN)} where the model parameters are assumed to be probabilistic. Figure (b) illustrates \textbf{MC dropout} where red denotes masked nodes and green denotes masked connections. With each variational inference the dropout configuration changes, introducing randomness into the neural network. Figure (c) illustrates \textbf{Epoch ensemble generation} where model parameters at each epoch are treated as independent ensemble members.}
    \vspace{-0.5cm}
    \label{fig:BNN}
\end{figure}

The BBB uncertainty estimation approach implemented in the Bayesian Transformer is compared to MC dropout and epoch ensemble generation. Figure \ref{fig:BNN} provides a theoretical comparison of the three epistemic uncertainty approaches. MC dropout is considered an efficient approximation of a BNN \cite{ABDAR2021243}, where randomness is introduced into the model by masking different nodes or connections with each inference (Figure \ref{fig:BNN}(b)). The dropout mask, $r$, follows a Bernoulli distribution as follows:

\begin{equation}
r \sim \text{Bernoulli}(p)
\label{eq_bern}
\end{equation}

where $p$ is the probability of a node or connection being maintained. Dropout is applied in the high-resolution Transformer architecture shown in Figure \ref{fig:Transformer}(a) for 10\% of nodes within the GloFormer and LoFormer blocks and in both MLP layers. VI with MC Dropout is repeated with different node configurations for each inference and the predictive mean and variance are calculated using Equations \ref{eq_mean} and \ref{eq_variance}, respectively.

Epoch ensemble generation considers the model parameters at each epoch as independent ensemble members, as outlined in Figure \ref{fig:BNN}(c). This method mitigates the cost of training multiple models while leveraging both high- and low-frequency patterns learned over the course of training. After each epoch, the current model weights are used to predict SIC. The mean and variance of all epoch predictions are calculated using Equations \ref{eq_mean} and \ref{eq_variance}, respectively.

\subsection{Decision-Level Data Fusion}
\begin{figure*}[]
        \centering
        \includegraphics[width = \textwidth]{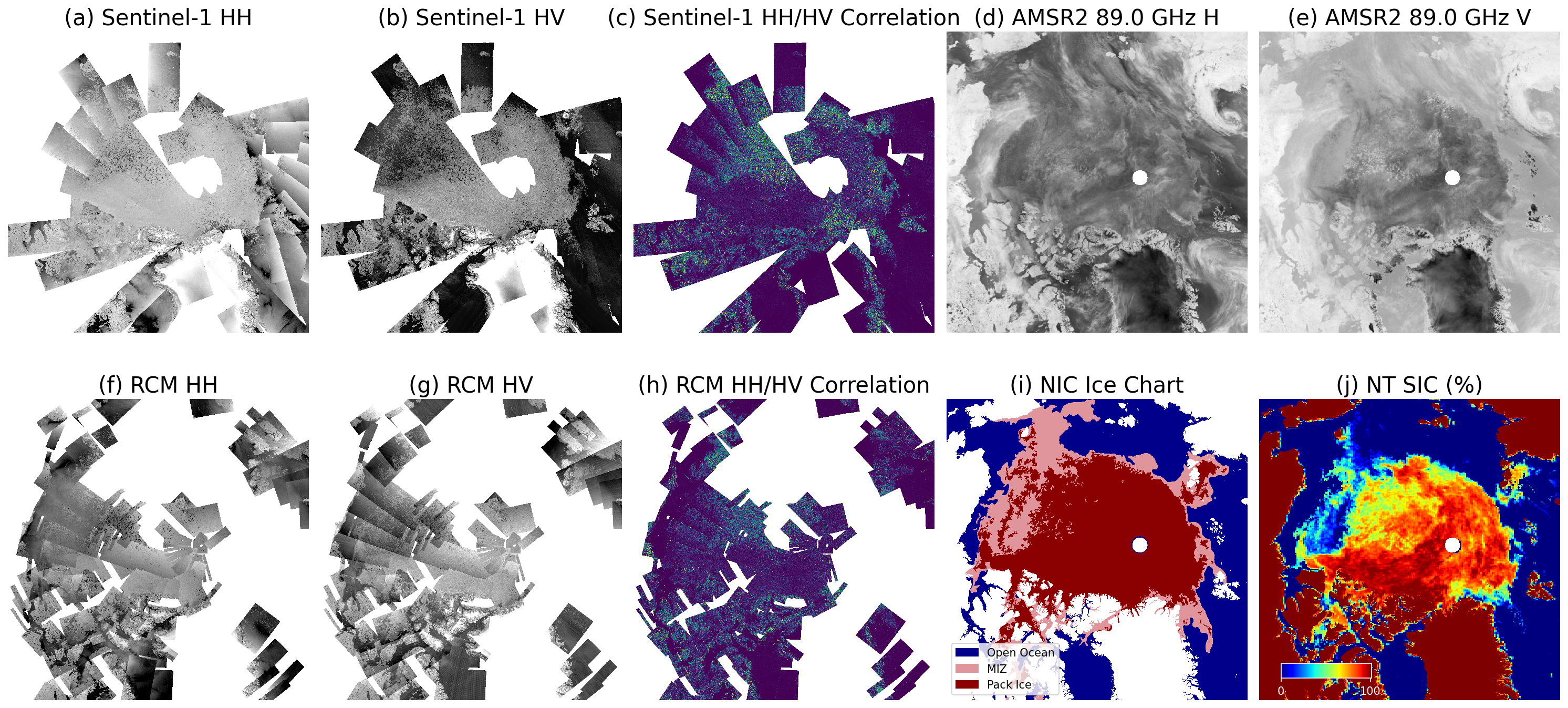}
        \vspace{-0.5cm}
        \caption{Overview of data downloaded and processed for September 4th, 2021. Includes Sentinel-1 (a) HH, (b) HV, and (c) HH and HV cross-polarization, RCM (f) HH, (g) HV, and (h) HH and HV cross-polarization, AMSR2 (d) 89.0 GHz H and (e) 89.0 GHz V data, as well as (i) NIC Ice Chart and (j) NASA Team SIC products.}
        \label{fig:data_maps}
\end{figure*}

Figure \ref{fig:Transformer}(d) outlines the decision-level data fusion scheme that is used to combine the Sentinel-1 ($S_{Sentinel-1}$), RCM ($S_{RCM}$), and 89.0GHz AMSR2-derived ($S_{AMSR2-89.0GHz}$) SIC into daily pan-Arctic maps ($\mathcal{M}_{fused}$). The fusion process is further detailed in Algorithm \ref{alg:sic_fusion}, which generates both the fused SIC estimate ($\mathcal{M}_{fused}$) and its corresponding uncertainty map ($\sigma_{fused}$). Consistency is maintained across all three datasets by using the same daily NASA Team SIC estimates to train the three models. 

The drawing order is determined according to the spatial resolution, sensor characteristics, and daily coverage inherent to each system. According to Table \ref{tab:data_spec}, Sentinel-1 EW mode has the highest spatial resolution, followed by RCM and AMSR2, respectively. Comparing SAR sensor characteristics, Sentinel-1 has standardized pre-processing tools that allow for incidence angle and noise corrections. Therefore, RCM is more prone to incidence angle and thermal banding than Sentinel-1, as shown in Figure \ref{fig:data_maps}(f) and (g). AMSR2 has complete daily coverage, yet the higher frequency 89.0 GHz channel is affected by atmospheric conditions. Given these factors, the chosen drawing order is to layer Sentinel-1 on top of RCM and AMSR2 SIC and uncertainty estimates, respectively.

\begin{algorithm}
\caption{Decision-Level Data Fusion for Pan-Arctic SIC and Uncertainty Mapping}
\label{alg:sic_fusion}
\begin{algorithmic}[1]
\State \textbf{Input:} $S_{Sentinel-1}, S_{RCM}, S_{AMSR2-89.0GHz}, $
\State \qquad \qquad $\sigma_{Sentinel-1}, \sigma_{RCM}, \sigma_{AMSR2-89.0GHz}$
\State \textbf{Output:} $\mathcal{M}_{fused}$, $\sigma_{fused}$
\State $\mathcal{M}_{fused} \leftarrow S_{AMSR2-89.0GHz}$
\State $\sigma_{fused} \leftarrow \sigma_{AMSR2-89.0GHz}$
\If{Availability($S_{RCM}$)}
    \State $\mathcal{M}_{fused} \leftarrow \text{Update}(\mathcal{M}_{fused}, S_{RCM})$ 
    \State $\sigma_{fused} \leftarrow \text{Update}(\sigma_{fused}, \sigma_{RCM})$ 
\EndIf
\If{Availability($S_{Sentinel-1}$)}
    \State $\mathcal{M}_{fused} \leftarrow \text{Update}(\mathcal{M}_{fused}, S_{Sentinel-1})$ 
    \State $\sigma_{fused} \leftarrow \text{Update}(\sigma_{fused}, \sigma_{Sentinel-1})$ 
\EndIf
\State
\Return $\mathcal{M}_{fused}$, $\sigma_{fused}$
\end{algorithmic}
\end{algorithm}

\subsection{Experimental Setup}
Extreme minimum sea ice conditions at the end of summer are a critical indicator of climate change \cite{wang_seasonal_2025}. However, accurately mapping SIC at the end of the summer season remains highly challenging due to the expanded MIZ, increased ambiguity in sea ice signatures, the presence of melt ponds that can result in underestimation of SIC, and a greater prevalence of small features such as leads and floes. Every year this annual minimum occurs in mid- to late September, marking the end of summer melt season and the beginning of winter freeze-up. 

Therefore, to leverage a comprehensive pan-Arctic dataset with complete spatial coverage of diverse ice types, including multiyear, first year, and young ice, this study concentrates on the pivotal pan-Arctic minimum-extent period in September. Sentinel-1, RCM, AMSR2, and NIC ice chart are downloaded, pre-processed, and divided into 256 x 256 image chips with 20\% overlap for 10 random dates in September 2021 and 3 random dates in September 2025. Table \ref{tab:sample_count} summarizes the temporal distribution of training, validation, and test samples. These years are chosen as they are the most recent years with two active Sentinel-1 satellites. Seven days are used for training (September 1st, 7th, 10th, 14th, 21st, 24th, and 30th, 2021), three days are used for validation (September 4th, 18th, and 27th, 2021), and three days are used for testing (September 3rd, 10th, and 30th, 2025). 

For each Sentinel-1, RCM, and AMSR2 dataset the deterministic and Bayesian High-Resolution Transformer models are trained for 50 epochs with learning rate 0.0001 and batch size 4. Figure \ref{fig:data_maps} provides an overview of the data downloaded for September 4th, 2021, which is the consistent across all dates. The best models are chosen during validation according to the lowest loss value. Variational inference is performed with 30 inferences to calculate the predictive mean and standard deviation for the probabilistic uncertainty estimation approaches. For epoch ensemble generation, the predictive mean and uncertainty are calculated from the predictions at each epoch. Finally, decision-level data fusion is performed by combining the predictive mean SIC maps and corresponding uncertainty derived from Sentinel-1, RCM, and AMSR2. SIC validation and test accuracies are assessed according to the NASA Team ground truth and ASI SIC products. All experiments were run on an AMD Ryzen Threadripper PRO 5975WX CPU with an NVIDIA RTX 6000 Ada GPU.

\begin{table}[!htbp]
\renewcommand{\arraystretch}{1.2}
\centering
\caption{Temporal Distribution of 256 x 256 Samples for Training, Validation, and Testing.}
\begin{tabular}{ll|l|l|l}
\toprule
\multicolumn{1}{l|}{Date}       & Dataset    & Sentinel-1 & RCM    & AMSR2  \\ \hline
\multicolumn{1}{l|}{2021/09/01} & Training   & 5858       & 4393   & 10,194 \\ \hline
\multicolumn{1}{l|}{2021/09/04} & Validation & 5270       & 4694   & 10,195 \\ \hline
\multicolumn{1}{l|}{2021/09/07} & Training   & 6333       & 3923   & 10,195 \\ \hline
\multicolumn{1}{l|}{2021/09/10} & Training   & 6058       & 4613   & 10,195 \\ \hline
\multicolumn{1}{l|}{2021/09/14} & Training   & 4612       & 2819   & 10,196 \\ \hline
\multicolumn{1}{l|}{2021/09/18} & Validation & 6568       & 8014   & 10,195 \\ \hline
\multicolumn{1}{l|}{2021/09/21} & Training   & 5502       & 7014   & 10,195 \\ \hline
\multicolumn{1}{l|}{2021/09/24} & Training & 5790       & 7428   & 10,195 \\ \hline
\multicolumn{1}{l|}{2021/09/27} & Validation   & 5784       & 7596   & 10,196 \\ \hline
\multicolumn{1}{l|}{2021/09/30} & Training   & 5911       & 7531   & 10,195 \\ \hline
\multicolumn{1}{l|}{2025/09/03} & Testing   & 5837       & 6893   & 10,205 \\ \hline
\multicolumn{1}{l|}{2025/09/10} & Testing   & 6245       & 6308   & 10,205 \\ \hline
\multicolumn{1}{l|}{2025/09/30} & Testing   & 5870       & 7449   & 10,205 \\ \hline
\multicolumn{2}{l|}{Total Training Samples}   & 40,064     & 37,721 & 71,365 \\ \hline
\multicolumn{2}{l|}{Total Validation Samples} & 17,622     & 20,304 & 30,586 \\ \hline
\multicolumn{2}{l|}{Total Testing Samples} & 17,952     & 20,650 & 30,615 \\
\bottomrule
\end{tabular}
\label{tab:sample_count}
\vspace{-0.3cm}
\end{table}

\subsection{Quality Metrics}

\subsubsection{Model Calibration}

Model calibration provides insight into the reliability of an uncertainty quantification approach by assessing whether predicted uncertainty corresponds to observed errors. To evaluate calibration, the weighted expected calibration error (ECE) \cite{naeini2015obtaining,wang_uncertainty_2025,wulf2024panarctic} is computed for MC dropout, epoch ensemble generation, and the Bayesian Transformer. For regression, the weighted ECE is adapted by standardizing residuals using the predicted standard deviation and converting the resulting z-scores into nominal two-sided confidence levels via the Normal cumulative distribution function. The standardized z-score for a single pixel, $z_i$, is calculated as follows: 

\begin{equation}
z_i = \frac{\left| y_i - \hat{y}_i \right|}{\sigma_i}
\end{equation}

where $y_i$ is the NASA Team SIC, $\hat{y}_i$ is the mean predicted SIC, and $\sigma_i$ is the mean predicted standard deviation. The nominal two-sided Gaussian confidence level for a single pixel, $p_i$, is defined as:

\begin{equation}
p_i = 2 \Phi(z_i) - 1
\end{equation}

where $\Phi(\cdot)$ is the cumulative distribution function of the standard normal distribution. These observed confidence levels are then binned and averaged to compare to their expected confidence levels, with each bin weighted by the sample distribution as follows: 

\begin{equation}
\mathrm{ECE}=\sum_{m=1}^{M}\frac{n_m}{N}\left|\bar{p}_m - p_m^{expected}\right|
\end{equation}

where $n_m$ is the number of pixels in bin $m$, $M$ is the total number of bins, $N$ is the total number of pixels, $\bar{p}_m$ is the mean observed confidence level, and $p_m^{expected}$ is the expected confidence level. This procedure evaluates whether the predicted uncertainty is statistically consistent with the distribution of residuals. A low ECE corresponds to a better calibrated model with more reliable uncertainty estimates.

\subsubsection{Feature Detection Accuracy}

\begin{figure}[]
    \centering
    \includegraphics[width=\linewidth]{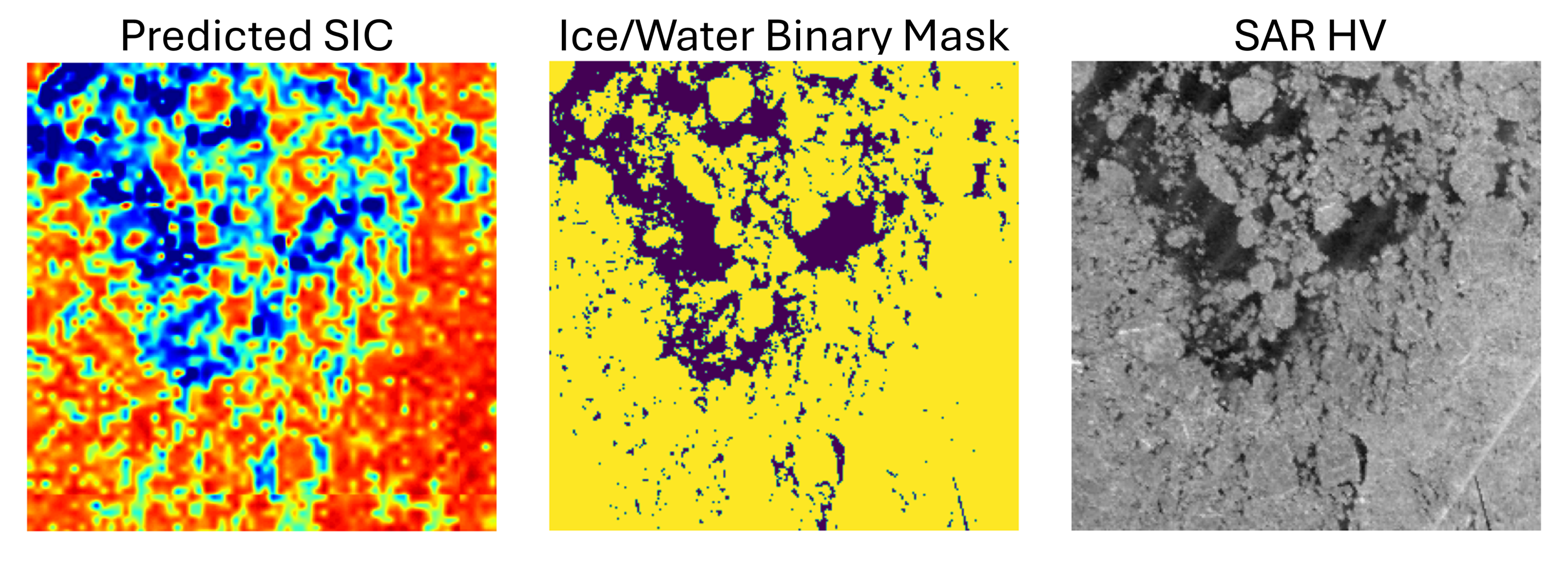}
    \vspace{-0.5cm}
    \caption{Comparison of predicted SIC (left), the ice/water binary ground truth mask derived using adaptive thresholding of the NIC ice chart (middle), and corresponding SAR HV (right) for one 256 x 256 validation sample.}
    \label{fig:ground_truth}
\end{figure}

Feature detection accuracy is assessed by manually selecting 300 high quality validation samples where adaptive thresholding of the NIC ice chart has achieved accurate ice and water binary masks, as shown in Figure \ref{fig:ground_truth}. The predicted SIC is converted to a predicted ice and water mask by setting a threshold of 15\% and 80\% in the MIZ and ice pack, respectively. The total number of true positives ($TP$), false positives ($FP$), true negatives ($TN$), and false negatives ($FN$) are used to derive feature detection metrics. For safe navigation, the most valuable metric is Ice Accuracy, or recall, calculated as follows:

\begin{equation}
\text{Ice Accuracy} = \frac{TP}{TP + FN}
\end{equation}

where $TP$ is the number of correctly predicted ice pixels and $TP + FN$ is the total number of actual ice pixels. This metric is essential for safe navigation because it quantifies the proportion of true ice pixels that are correctly detected, ensuring hazardous ice features are not missed in operational mapping. To ensure the model is not hallucinating ice, the water accuracy, or sensitivity, is also calculated:

\begin{equation}
\text{Water Accuracy} = \frac{TN}{TN + FP}
\end{equation}

where $TN$ is the number of correctly predicted water pixels and $TN + FP$ is the total number of actual water pixels. Finally, balanced overall accuracy is computed to provide a single summary measure of feature detection performance under class imbalance.

\begin{equation}
\text{Overall Accuracy} =\frac{\text{Ice Accuracy}+\text{Water Accuracy}}{2}
\end{equation}

For all three feature detection metrics a higher score indicates greater accuracy when detecting sea ice features. 

\subsubsection{Pan-Arctic SIC Accuracy}
Pan-Arctic SIC accuracy is evaluated against the University of Bremen 3125 m ASI product for the 2021 validation and 2025 test datasets. Averaging is used to downsample high-resolution SIC predictions to the resolution of the corresponding ASI product. SIC quality is assessed according to R\textsuperscript{2}, mean absolute error (MAE), and mean error (ME), calculated as follows:

\begin{equation}
R^2 = 1 - \frac{\sum_{i=1}^{n} (y_i - \hat{y}_i)^2}{\sum_{i=1}^{n} (y_i - \bar{y})^2}
\end{equation}

\begin{equation}
\text{MAE} = \frac{1}{n} \sum_{i=1}^{n} \left| y_i - \hat{y}_i \right|
\end{equation}

\begin{equation}
\text{ME} = \frac{1}{n} \sum_{i=1}^{n} \left( y_i - \hat{y}_i \right)
\end{equation}

where $y_i$ is the expected SIC, $\hat{y}_i$ is the predicted SIC, $\bar{y}$ is the mean expected SIC, and $n$ is the total number of pixels. R\textsuperscript{2} measures the proportion of variance explained by the model, MAE quantifies the average magnitude of errors regardless of direction, and ME indicates whether predictions systematically over- or under-estimate SIC. A high R\textsuperscript{2} score along with MAE and ME scores near 0 indicate better pan-Arctic SIC accuracy. Together these metrics evaluate both model performance and systematic bias.

\section{Results and Analysis}
\label{experiments}

\subsection{Pan-Arctic SIC and Uncertainty Visualization} 

\begin{figure*}[htbp]
     \centering
     % First Subfigure
     \begin{subfigure}[b]{\textwidth}
         \centering
         \includegraphics[width=\textwidth]{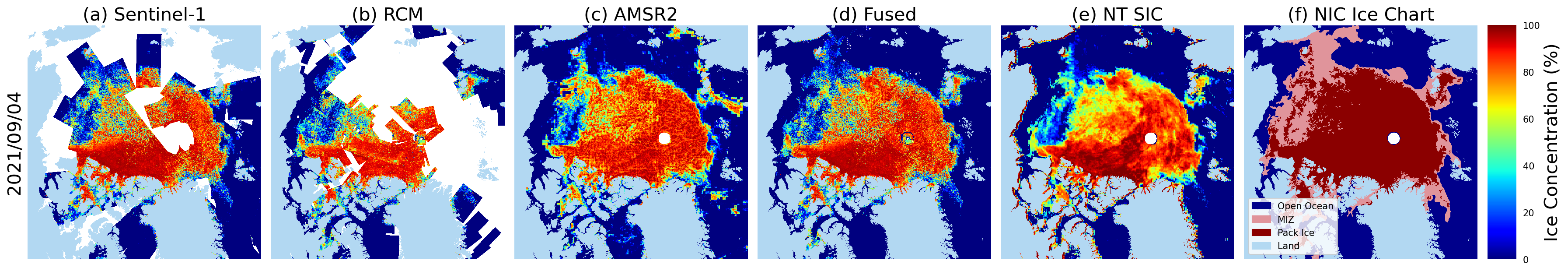}
     \end{subfigure}
     
     % Second Subfigure
     \begin{subfigure}[b]{\textwidth}
         \centering
         \includegraphics[width=\textwidth]{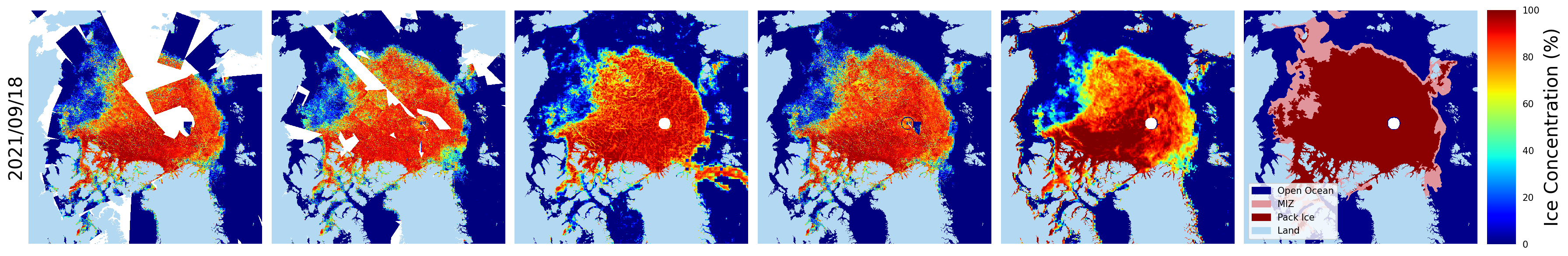}
     \end{subfigure}
     
    \begin{subfigure}[b]{\textwidth}
         \centering
         \includegraphics[width=\textwidth]{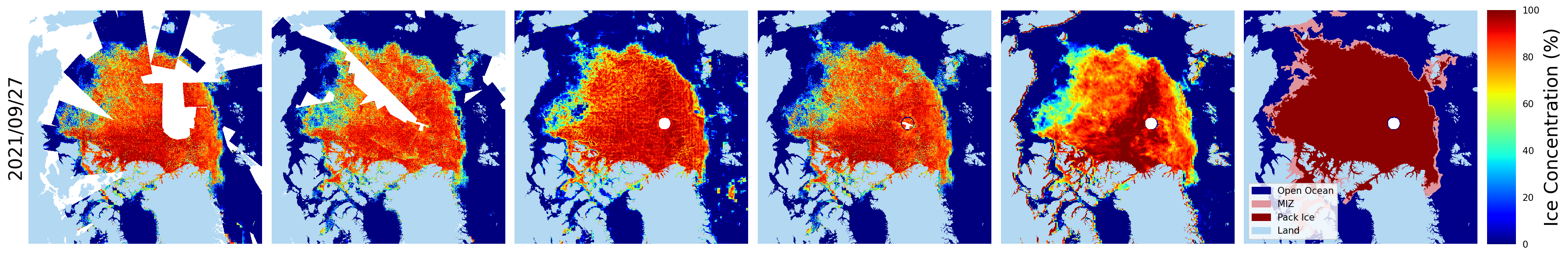}
     \end{subfigure}
     
     \caption{Pan-Arctic SIC from the Bayesian High-Resolution Transformer for \textbf{validation data} on September 4th, 18th, and 27th, 2021. SIC derived from (a) Sentinel-1, (b) RCM, and (c) AMSR2, and (d) is the final fused map of all data sources with Sentinel-1 layered on top, followed by RCM and AMSR2. SIC estimates are visually compared to (e) NASA Team SIC and (f) NIC Ice Chart.}
     \label{fig:ic_maps}

     \centering
     % First Subfigure
     \begin{subfigure}[b]{\textwidth}
         \centering
         \includegraphics[width=\textwidth]{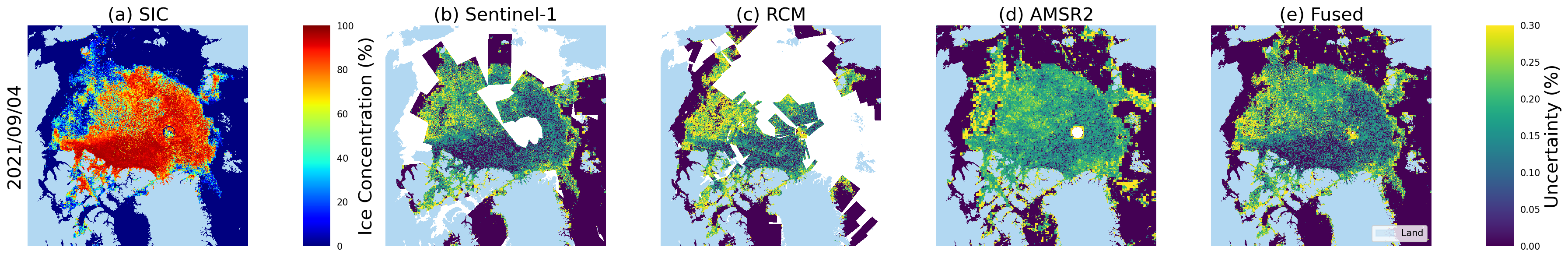}
     \end{subfigure}
     
     % Second Subfigure
     \begin{subfigure}[b]{\textwidth}
         \centering
         \includegraphics[width=\textwidth]{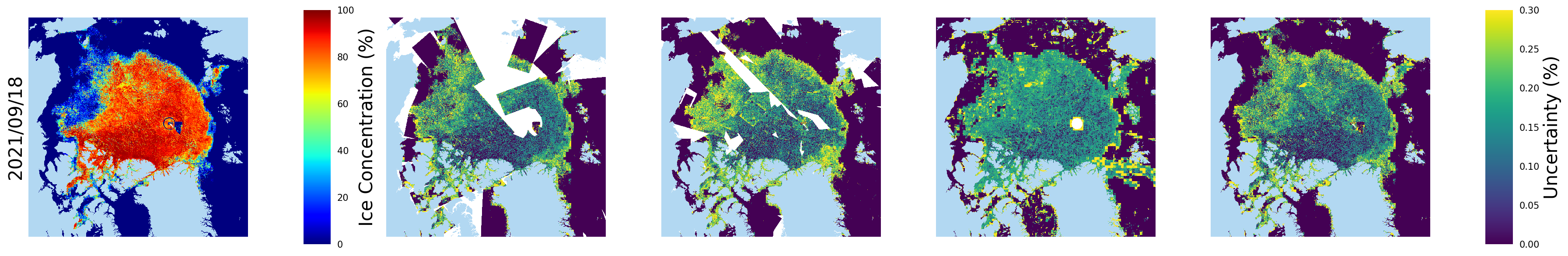}
     \end{subfigure}
     
    \begin{subfigure}[b]{\textwidth}
         \centering
         \includegraphics[width=\textwidth]{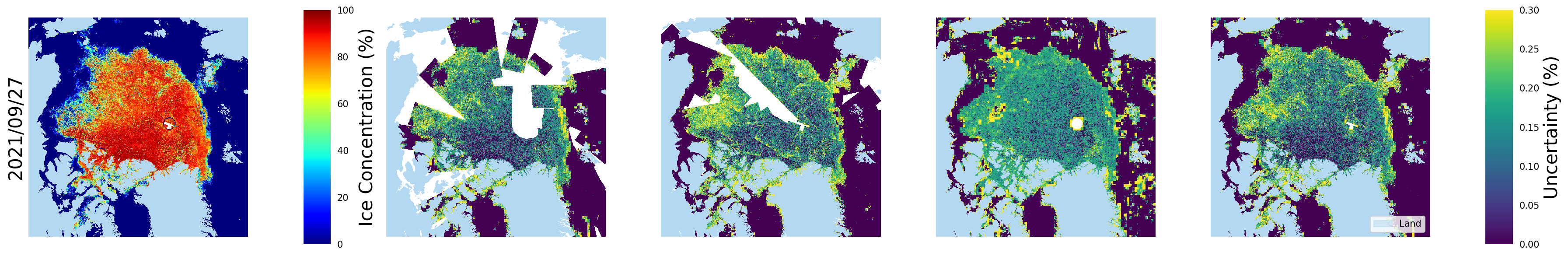}
     \end{subfigure}
     
     \caption{(a) Fused Pan-Arctic SIC map and corresponding uncertainty from the Bayesian High-Resolution Transformer for \textbf{validation data} on September 4th, 18th, and 27th, 2021. SIC and uncertainty are derived from (b) Sentinel-1, (c) RCM, and (d) AMSR2, and (e) is the final fused uncertainty map of all data sources with Sentinel-1 layered on top, followed by RCM and AMSR2.}
     \label{fig:unc_maps}
\end{figure*}

\begin{figure*}[htbp]
     \centering
     % First Subfigure
     \begin{subfigure}[b]{\textwidth}
         \centering
         \includegraphics[width=\textwidth]{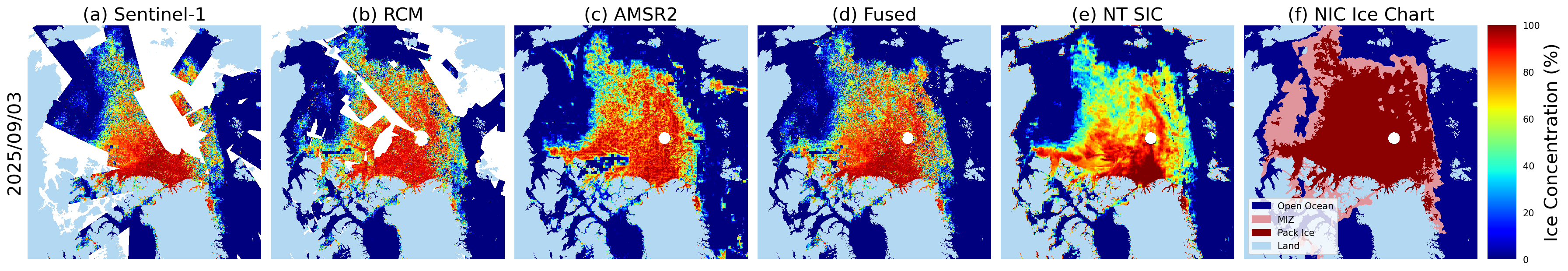}
     \end{subfigure}
     
     % Second Subfigure
     \begin{subfigure}[b]{\textwidth}
         \centering
         \includegraphics[width=\textwidth]{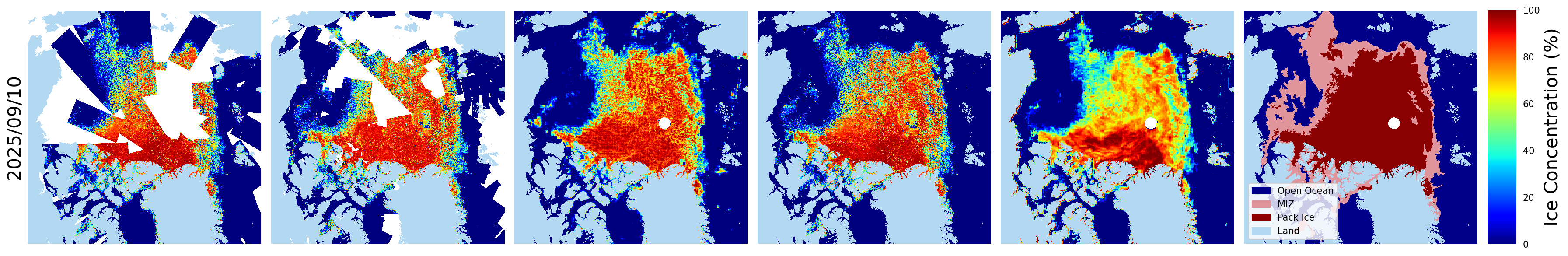}
     \end{subfigure}
     
    \begin{subfigure}[b]{\textwidth}
         \centering
         \includegraphics[width=\textwidth]{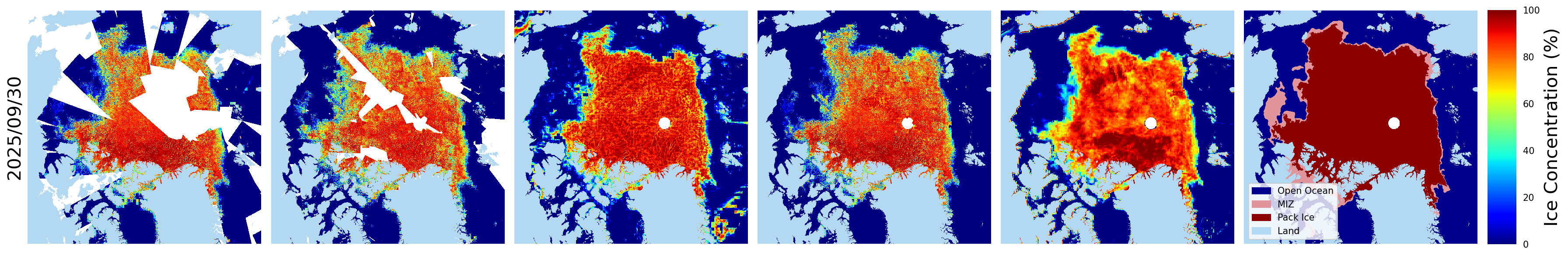}
     \end{subfigure}
     
     \caption{Pan-Arctic SIC from the Bayesian High-Resolution Transformer for \textbf{test data} on September 3rd, 10th, and 30th, 2025. SIC derived from (a) Sentinel-1, (b) RCM, and (c) AMSR2, and (d) is the final fused map of all data sources with Sentinel-1 layered on top, followed by RCM and AMSR2. SIC estimates are visually compared to (e) NASA Team SIC and (f) NIC Ice Chart.}
     \label{fig:ic_maps_test}

     \centering
     % First Subfigure
     \begin{subfigure}[b]{\textwidth}
         \centering
         \includegraphics[width=\textwidth]{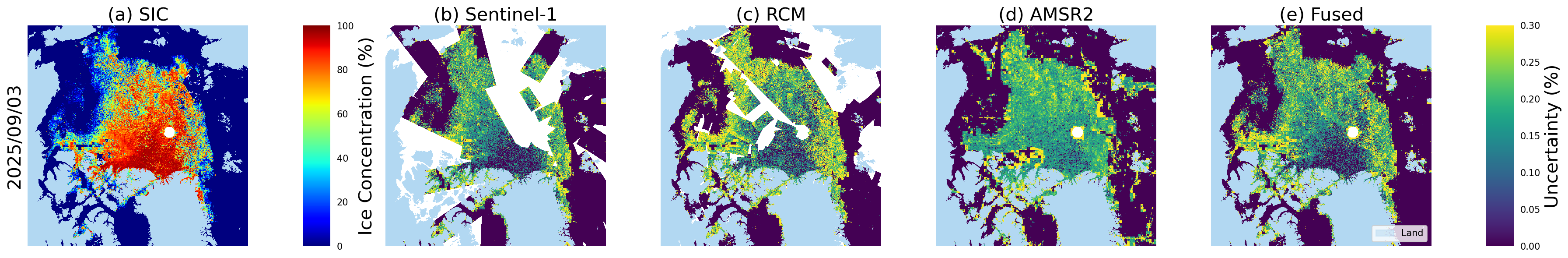}
     \end{subfigure}
     
     % Second Subfigure
     \begin{subfigure}[b]{\textwidth}
         \centering
         \includegraphics[width=\textwidth]{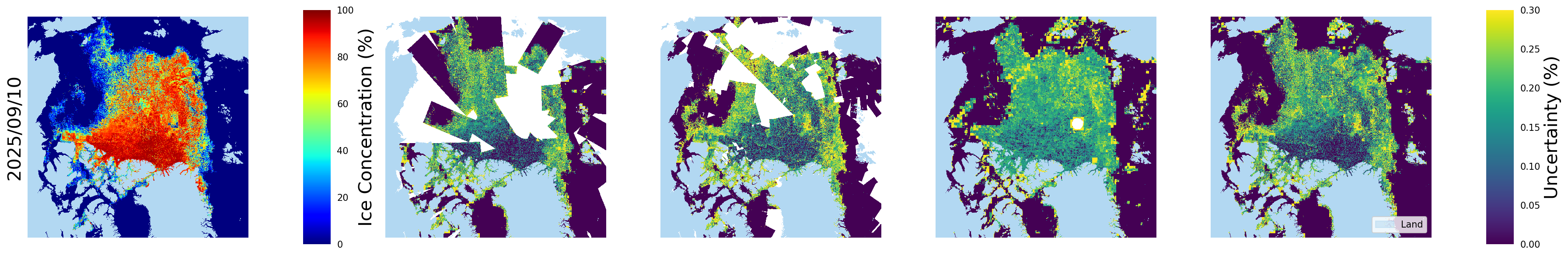}
     \end{subfigure}
     
    \begin{subfigure}[b]{\textwidth}
         \centering
         \includegraphics[width=\textwidth]{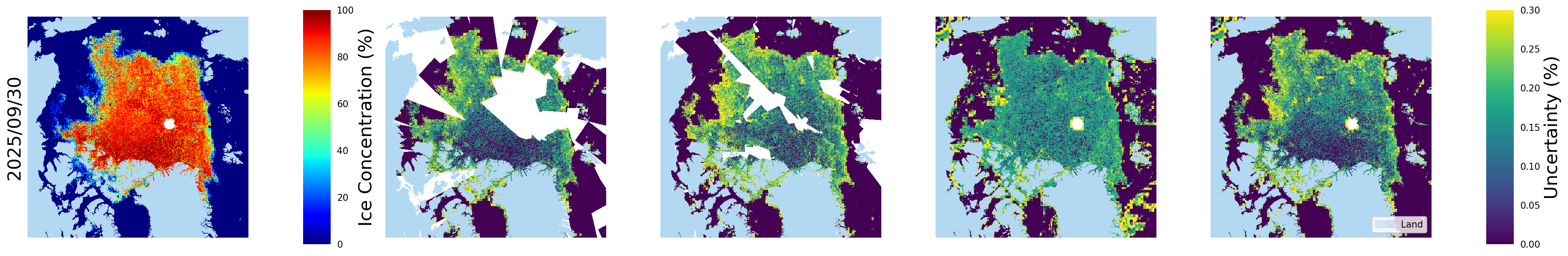}
     \end{subfigure}
     
     \caption{(a) Fused Pan-Arctic SIC map and corresponding uncertainty from the Bayesian High-Resolution Transformer for \textbf{test data} on September 3rd, 10th, and 30th, 2025. SIC and uncertainty are derived from (b) Sentinel-1, (c) RCM, and (d) AMSR2, and (e) is the final fused uncertainty map of all data sources with Sentinel-1 layered on top, followed by RCM and AMSR2.}
     \label{fig:unc_maps_test}
\end{figure*}

Figure \ref{fig:ic_maps} presents the pan-Arctic SIC maps generated by the Bayesian High-Resolution Transformer for the validation dataset on September 4th, 18th, and 27th, 2021, with the associated uncertainties in Figure \ref{fig:unc_maps}. Similar results for the test dataset on September 3rd, 10th, and 30th, 2025 are presented in Figures \ref{fig:ic_maps_test} and \ref{fig:unc_maps_test}, respectively. Comparing the fused pan-Arctic SIC map in Figure \ref{fig:ic_maps}(d) to the NASA Team SIC product in Figure \ref{fig:ic_maps}(e), the Bayesian Transformer achieved a more detailed representation while preserving the large-scale SIC patterns. This fused map also shows greater agreement with the NIC ice chart, particularly in the MIZ. This can be attributed to the integration of SAR systems, as both the Bayesian Transformer SIC and the NIC ice charts are informed by these datasets. These findings are consistent with the test results in Figure \ref{fig:ic_maps_test}, which presents a more expansive MIZ than the validation data in 2021.

Across all data types, Figures \ref{fig:unc_maps} and \ref{fig:unc_maps_test} demonstrate that the proposed Bayesian Transformer model was more uncertain in the MIZ than in open water or ice pack, which is supported by previous literature \cite{chen2023predicting}. The MIZ is highly dynamic and changes rapidly, and its subtle sea ice signatures in SAR imagery can easily be mistaken for wind effects or noise over open water. Errors in the AMSR2-derived SIC over open water for September 18th and 27th can be attributed to the atmospheric noise in the 89.0 GHz AMSR2 channel. The corresponding higher uncertainty estimates in these region show that the model was not confident in these predictions. Therefore, these results support our choice to trust SAR more than the 89.0 GHz channel when creating pan-Arctic SIC maps.

\subsection{Uncertainty Estimation Comparison}

\begin{table}[!htbp]
\renewcommand{\arraystretch}{1.2}
\centering
\caption{Weighted Expected Calibration Error (ECE) for MIZ and Ice Pack. Best performance is \textbf{bolded}.}
\begin{tabular}{c|c|c|c}
\toprule
Dataset                     & Method               & MIZ & Ice Pack \\ \hline
\multirow{3}{*}{Sentinel-1} & MC Dropout           & 0.0171  & 0.0226       \\
                            & Epoch Ensemble       & 0.0169  & 0.0207       \\
                            & \textbf{Bayesian Transformer} & \textbf{0.0018}  & \textbf{0.0022}       \\ \hline
\multirow{3}{*}{RCM}        & MC Dropout           & 0.0145  & 0.0238       \\
                            & Epoch Ensemble       & 0.0144  & 0.0194       \\
                            & \textbf{Bayesian Transformer} & \textbf{0.0019}  & \textbf{0.0017}       \\ \hline
\multirow{3}{*}{AMSR2}      & MC Dropout           & 0.0059  & 0.0066       \\
                            & Epoch Ensemble       & 0.0264  & 0.0299       \\
                            & \textbf{Bayesian Transformer} & \textbf{0.0019}  & \textbf{0.0028}      \\
\bottomrule
\end{tabular}
\label{tab:calibration_tab}
%\vspace{-0.3cm}
\end{table}

\begin{figure*}[!htbp]
    \centering
    \includegraphics[width=\linewidth]{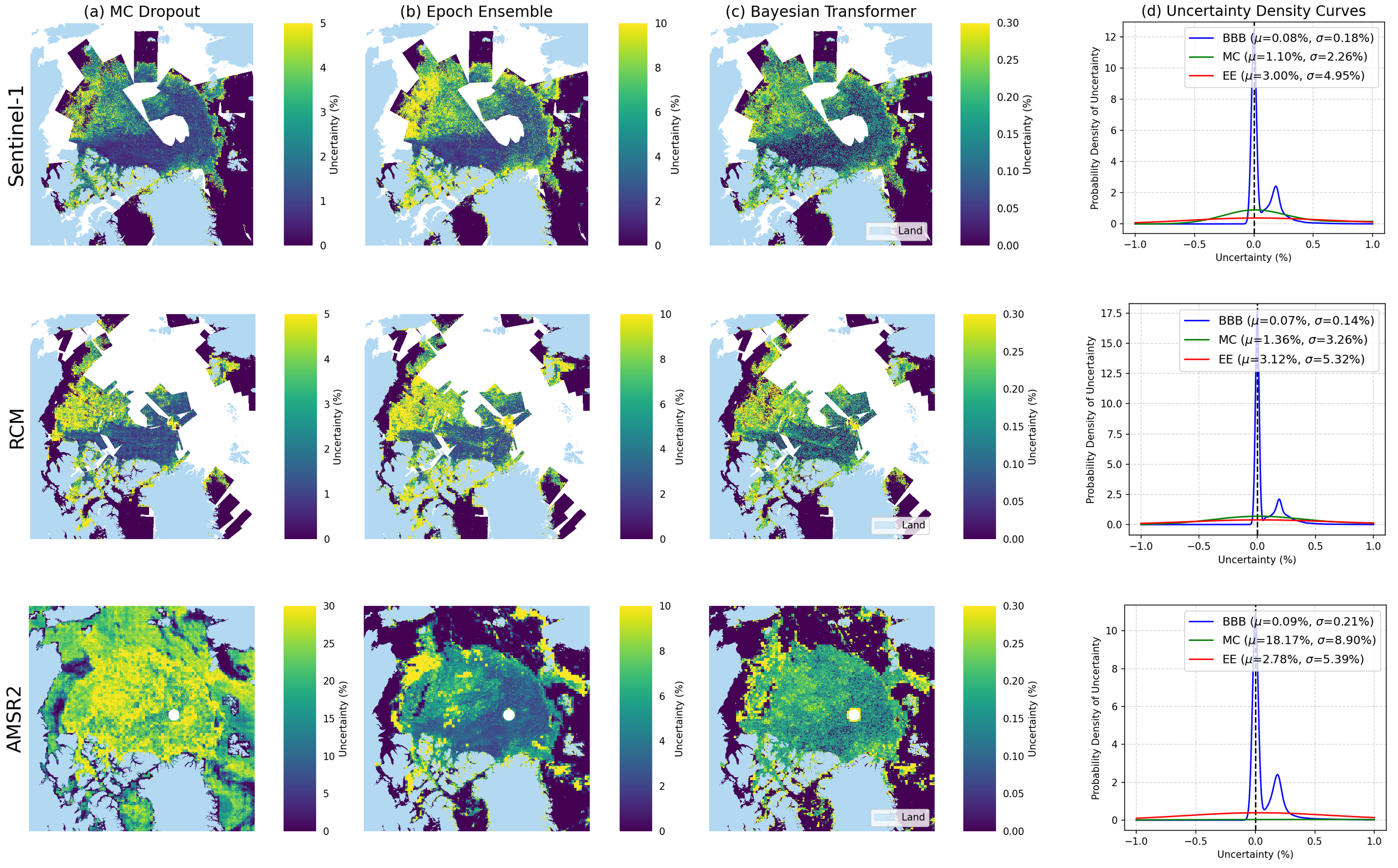}
    \vspace{-0.5cm}
    \caption{Comparison of uncertainty quantification approaches for Sentinel-1, RCM, and AMSR2. Uncertainty estimates are shown using (a) MC Dropout, (b) Epoch Ensemble, and (c) BBB implemented in the Bayesian Transformer for September 4th, 2021. The distribution of uncertainty for all validation samples are shown using (d) Gaussian Kernel Density curves for each dataset with corresponding mean and standard deviation.}
    \label{fig:uncert_compare}
\end{figure*}

\begin{table*}[!htbp]
\renewcommand{\arraystretch}{1.2}
\centering
\caption{Mean Uncertainty (\%) per Ice Concentration Class Achieved by Different Methods for Validation Data. Lowest overall uncertainty estimates are \textbf{bolded}.}
\begin{tabular}{c|c|llllllllll}
\toprule
                                           &                                & \multicolumn{10}{c}{Ice   Concentration Classes (\%)}                                                                                                                                                                                                                                                                                                                                                                                                                                                            \\ \cline{3-12} 
\multirow{-2}{*}{Uncertainty   Estimation} & \multirow{-2}{*}{Input   Data} & \multicolumn{1}{c|}{0-10}                         & \multicolumn{1}{c|}{10-20}                        & \multicolumn{1}{c|}{20-30}                        & \multicolumn{1}{c|}{30-40}                        & \multicolumn{1}{c|}{40-50}                        & \multicolumn{1}{c|}{50-60}                        & \multicolumn{1}{c|}{60-70}                        & \multicolumn{1}{c|}{70-80}                        & \multicolumn{1}{c|}{80-90}                        & \multicolumn{1}{c}{90-100}   \\ \hline
                                           & Sentinel-1                     & \multicolumn{1}{l|}{0.17}                         & \multicolumn{1}{l|}{5.38}                         & \multicolumn{1}{l|}{5.59}                         & \multicolumn{1}{l|}{5.35}                         & \multicolumn{1}{l|}{4.85}                         & \multicolumn{1}{l|}{4.37}                         & \multicolumn{1}{l|}{3.68}                         & \multicolumn{1}{l|}{2.65}                         & \multicolumn{1}{l|}{1.44}                         & 0.82                         \\
                                           & RCM                            & \multicolumn{1}{l|}{0.18}                         & \multicolumn{1}{l|}{7.53}                         & \multicolumn{1}{l|}{7.94}                         & \multicolumn{1}{l|}{8.94}                         & \multicolumn{1}{l|}{8.86}                         & \multicolumn{1}{l|}{7.80}                         & \multicolumn{1}{l|}{6.15}                         & \multicolumn{1}{l|}{4.13}                         & \multicolumn{1}{l|}{2.07}                         & 1.05                         \\
\multirow{-3}{*}{MC Dropout}    & AMSR2                          & \multicolumn{1}{l|}{7.95}                         & \multicolumn{1}{l|}{18.29}                        & \multicolumn{1}{l|}{22.57}                        & \multicolumn{1}{l|}{25.24}                        & \multicolumn{1}{l|}{28.04}                        & \multicolumn{1}{l|}{29.66}                        & \multicolumn{1}{l|}{27.15}                        & \multicolumn{1}{l|}{21.55}                        & \multicolumn{1}{l|}{13.81}                        & 4.49                         \\ \hline
                                           & Sentinel-1                     & \multicolumn{1}{l|}{0.72}                         & \multicolumn{1}{l|}{11.69}                        & \multicolumn{1}{l|}{12.65}                        & \multicolumn{1}{l|}{12.96}                        & \multicolumn{1}{l|}{12.52}                        & \multicolumn{1}{l|}{11.37}                        & \multicolumn{1}{l|}{9.49}                         & \multicolumn{1}{l|}{6.85}                         & \multicolumn{1}{l|}{3.80}                         & 1.94                         \\
                                           & RCM                            & \multicolumn{1}{l|}{0.64}                         & \multicolumn{1}{l|}{10.64}                        & \multicolumn{1}{l|}{12.31}                        & \multicolumn{1}{l|}{14.03}                        & \multicolumn{1}{l|}{14.29}                        & \multicolumn{1}{l|}{13.32}                        & \multicolumn{1}{l|}{11.45}                        & \multicolumn{1}{l|}{8.54}                         & \multicolumn{1}{l|}{4.45}                         & 2.45                         \\
\multirow{-3}{*}{Epoch   Ensemble}         & AMSR2                          & \multicolumn{1}{l|}{0.86}                         & \multicolumn{1}{l|}{11.37}                        & \multicolumn{1}{l|}{12.68}                        & \multicolumn{1}{l|}{13.82}                        & \multicolumn{1}{l|}{13.82}                        & \multicolumn{1}{l|}{12.46}                        & \multicolumn{1}{l|}{10.07}                         & \multicolumn{1}{l|}{6.63}                         & \multicolumn{1}{l|}{3.47}                         & 2.16                         \\ \hline
                                           & \textbf{Sentinel-1}                     & \multicolumn{1}{l|}{\textbf{0.01}} & \multicolumn{1}{l|}{\textbf{0.30}} & \multicolumn{1}{l|}{\textbf{0.32}} & \multicolumn{1}{l|}{\textbf{0.33}} & \multicolumn{1}{l|}{\textbf{0.32}} & \multicolumn{1}{l|}{\textbf{0.29}} & \multicolumn{1}{l|}{\textbf{0.24}} & \multicolumn{1}{l|}{\textbf{0.19}} & \multicolumn{1}{l|}{\textbf{0.13}} & \textbf{0.07} \\
                                           & RCM                            & \multicolumn{1}{l|}{0.01}                         & \multicolumn{1}{l|}{0.29}                         & \multicolumn{1}{l|}{0.34}                         & \multicolumn{1}{l|}{0.35}                         & \multicolumn{1}{l|}{0.34}                         & \multicolumn{1}{l|}{0.31}                         & \multicolumn{1}{l|}{0.26}                         & \multicolumn{1}{l|}{0.20}                         & \multicolumn{1}{l|}{0.13}                         & 0.08                         \\
\multirow{-3}{*}{\textbf{Bayesian Transformer}}   & AMSR2                          & \multicolumn{1}{l|}{0.02}                         & \multicolumn{1}{l|}{0.39}                         & \multicolumn{1}{l|}{0.38}                         & \multicolumn{1}{l|}{0.39}                         & \multicolumn{1}{l|}{0.36}                         & \multicolumn{1}{l|}{0.31}                         & \multicolumn{1}{l|}{0.26}                         & \multicolumn{1}{l|}{0.22}                         & \multicolumn{1}{l|}{0.18}                         & 0.12                \\
\bottomrule
\end{tabular}
\label{tab:uncert_tab}
\end{table*}

 The Bayesian Transformer achieved the lowest ECE across datasets, followed by epoch ensemble and MC dropout, as shown in Table \ref{tab:calibration_tab}. These findings are consistent with the uncertainty quantification comparison for lithological mapping \cite{wang_uncertainty_2025}, where the Bayesian CNN also had the lowest calibration error among MC dropout and ensemble generation. Additionally, the MIZ achieved lower calibration error than ice pack for most methods, save for the RCM Bayesian Transformer. This indicates that MIZ uncertainty is primarily driven by aleatoric variability arising from signature ambiguity and heterogeneous ice conditions, whereas in the homogeneous ice pack the errors were more attributed to systematic model bias. Therefore, the ECE can better explain the stochastic variability in the MIZ.

Each approach was also assessed according to spatial variation and distribution of uncertainty estimates. Figure \ref{fig:uncert_compare} compares the SIC uncertainty maps for each method for September 4th, 2021 along with the kernel density curves of uncertainty for the validation dataset. To assess uncertainty ranges across different ice types, Table \ref{tab:uncert_tab} compares the mean uncertainty per SIC class for each method. Overall, Table \ref{tab:uncert_tab} shows consistent conclusions with the uncertainty maps in Figures \ref{fig:unc_maps}, \ref{fig:unc_maps_test}, and \ref{fig:uncert_compare}. The MIZ (10\% to 80\%) had greater uncertainty than open water (0\% to 10\%) and ice pack (80\% to 100\%) for all methods and data sources. However, the Bayesian Transformer was the only method that maintained consistent uncertainty ranges and distributions for Sentinel-1, RCM, and AMSR2. Table \ref{tab:uncert_tab} demonstrates how the proposed Bayesian Transformer provides smaller and more consistent model uncertainties (0.01\% - 0.39\%) across data sources than MC dropout (0.17\% - 29.66\%) and epoch ensemble (0.64\% - 14.29\%). The density curves in Figure \ref{fig:uncert_compare}(d) support these results, where the Bayesian Transformer uncertainty curves consistently peaked twice at 0\% and approximately 0.2\% for all datasets. Although the greater variation in MC dropout and epoch ensemble uncertainty estimates may be more interpretable, the consistency in Bayesian Transformer across data types supports effective decision-level fusion for pan-Arctic SIC mapping.

When comparing uncertainty across different datasets, the AMSR2 and RCM-derived SIC had greater uncertainty than Sentinel-1-derived SIC for all methods, likely due to the cloud and fog that impacts the 89.0 GHz AMSR2 channel and the thermal noise and incidence angle effect more common in RCM. Overall, the proposed Bayesian Transformer produced the lowest model uncertainty with Sentinel-1, as shown in the bolded Table \ref{tab:uncert_tab} row. Therefore, our choice to place the AMSR2 and RCM SIC below the Sentinel-1 SIC during fusion is further supported by the uncertainties given by the Bayesian Transformer model. The least informative uncertainty map was produced using MC dropout for AMSR2-derived SIC, as shown in Figure \ref{fig:uncert_compare}(a). The uncertainty estimates were consistent across the entire pan-Arctic region without any patterns correlated with ice type. This is likely due to MC dropout having poor calibration, making it unable to interpret the atmospheric conditions in 89.0 GHz AMSR2. However, the low corresponding ECE in Table \ref{tab:calibration_tab} is misleading. This vanishing ECE would be caused by homogeneous SIC estimates, where all samples are binned together during the ECE calculation \cite{wulf2024panarctic}. Therefore, it is important to consider more than just the ECE when assessing uncertainty estimation performance.

As the Bayesian Transformer integrates uncertainty estimation into the training objective, it can produce uncertainty estimates that are more consistent and better calibrated across heterogeneous data than MC dropout and epoch ensemble generation. In addition, the Bayesian Transformer can mitigate variance as its Bayesian prior acts as a regularizer. The MC dropout approach is a simple approximation of BNN, allowing for simpler and more efficient uncertainty quantification. Yet, miscalibration in MC dropout results in poor performance when atmospheric interference and noise are more dominant in the data, as shown in the Figure \ref{fig:uncert_compare}(a) AMSR2 uncertainty map. Epoch ensemble generation is better calibrated than MC dropout without the extensive hyperparameter tuning required by probabilistic methods. Tuning the percentage of MC dropout or the choice of Bayesian prior and initialization of model parameters can be a difficult task and may be dependent on the unique characteristics of each dataset. However, it is still a heuristic interpretation of the true posterior distribution. The Bayesian Transformer thus offers more reliable uncertainty quantification for SIC mapping using heterogeneous datasets compared to other common approaches.

\subsection{Local Feature Detection Accuracy Assessment}
\label{sec:local_acc}
\subsubsection{Uncertainty Estimation Comparison}

\begin{table}[]
\renewcommand{\arraystretch}{1.2}
\centering
\caption{Uncertainty Estimation Comparison Overall Accuracy/Ice Accuracy/Water Accuracy for Feature Detection. Best performance is \textbf{bolded}.}
\begin{tabular}{c|c|c}
\toprule
\begin{tabular}[c]{@{}c@{}}Method\end{tabular} & Sentinel-1          & RCM                 \\ \hline
NASA Team                                                              & 0.55/0.81/0.28 & 0.55/0.83/0.26 \\ \hline
\begin{tabular}[c]{@{}c@{}}Deterministic\\ Transformer\end{tabular}     & 0.70/0.83/\textbf{0.57} & \textbf{0.69}/0.90/\textbf{0.48} \\ \hline
\begin{tabular}[c]{@{}c@{}}MC Dropout\end{tabular}     & 0.70/0.84/0.55 & 0.69/0.91/0.47 \\ \hline
\begin{tabular}[c]{@{}c@{}}Epoch\\ Ensemble\end{tabular}          & 0.70/0.86/0.53 & 0.59/0.85/0.33 \\ \hline
\begin{tabular}[c]{@{}c@{}}\textbf{Bayesian}\\ \textbf{Transformer}\end{tabular}          & \textbf{0.70}/\textbf{0.90}/0.52 & 0.66/\textbf{0.92}/0.40 \\
\bottomrule
\end{tabular}
\label{tab:feature_acc}
\vspace{-0.3cm}
\end{table}

\begin{figure*}[!t]
    \centering
    \includegraphics[width=\linewidth]{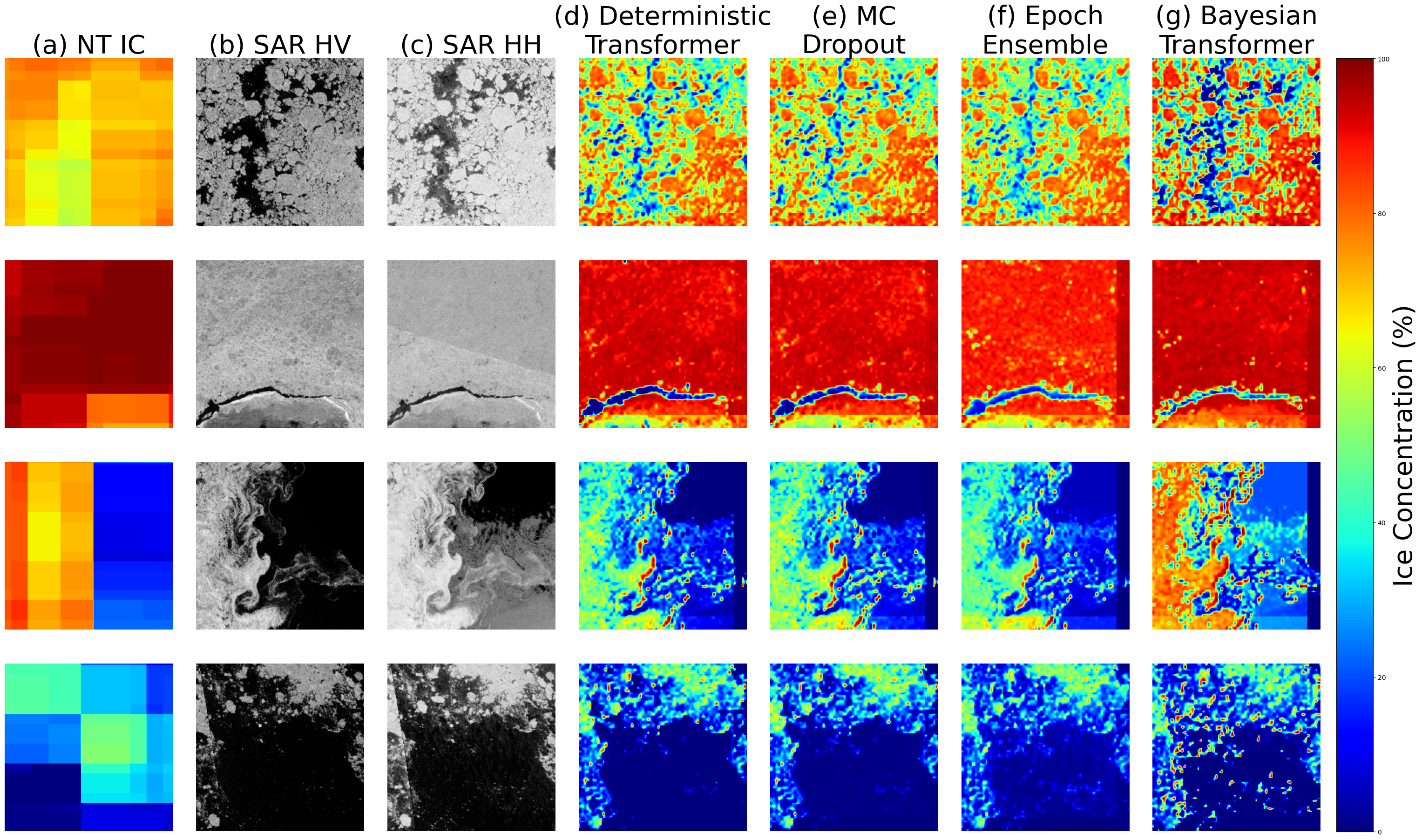}
    \vspace{-0.2cm}
    \caption{Local visual comparison of SIC derived from Sentinel-1 on September 4th, 2021, where (a) NASA Team SIC, (b) Sentinel-1 HV Imagery, (c) Sentinel-1 HH Imagery, (d) Deterministic Transformer SIC, (e) Mean Monte Carlo Dropout SIC, (f) Mean Epoch Ensemble SIC, and \textbf{(g) Mean Bayesian Transformer SIC (our approach)}.}
    \label{fig:S1_compare}
    \includegraphics[width=\linewidth]{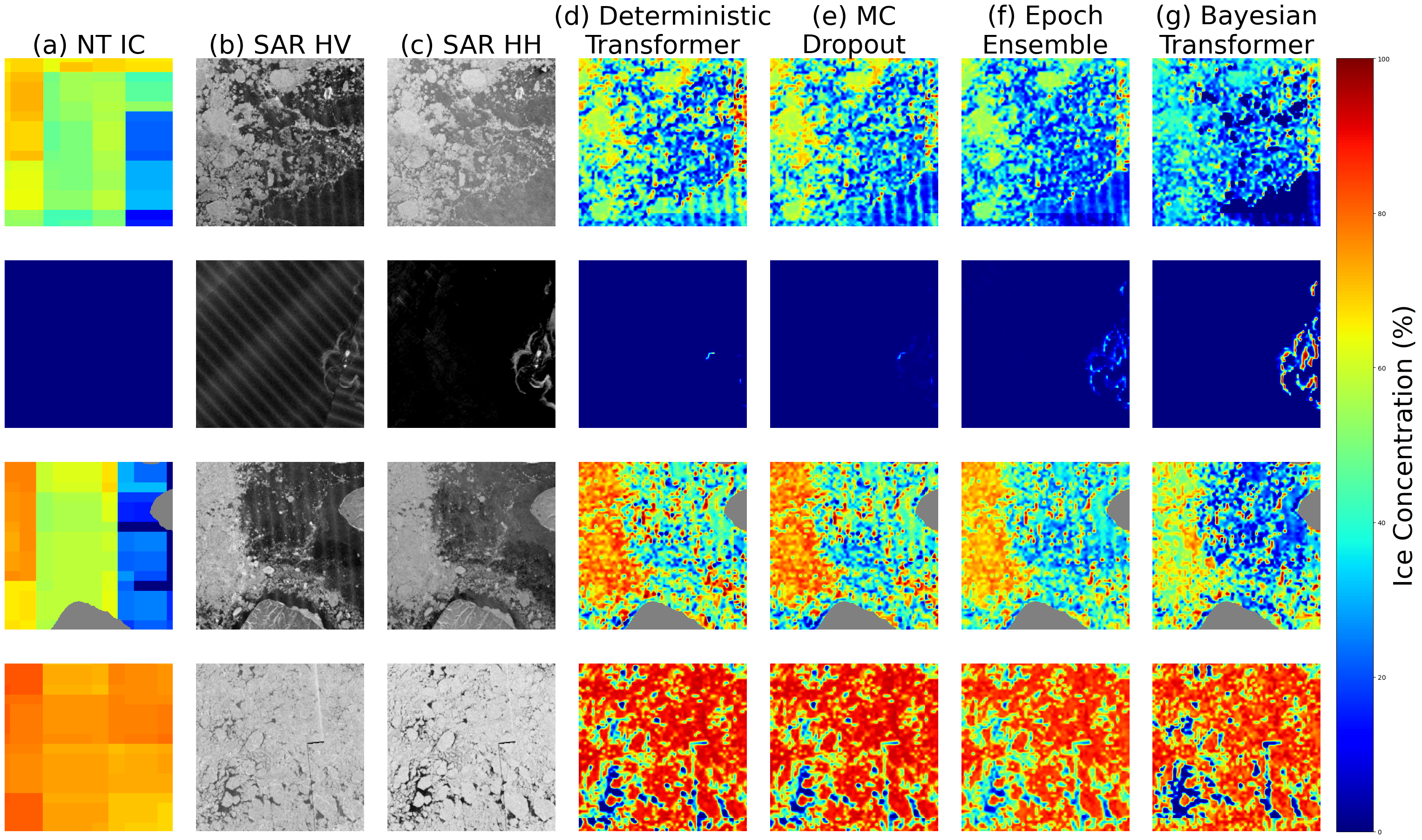}
    \vspace{-0.2cm}
    \caption{Local visual comparison of SIC derived from RCM on September 4th, 2021, where (a) NASA Team SIC, (b) RCM HV Imagery, (c) RCM HH Imagery, (d) Deterministic Transformer SIC, (e) Mean Monte Carlo Dropout SIC, (f) Mean Epoch Ensemble SIC, and \textbf{(g) Mean Bayesian Transformer SIC (our approach)}.}
    \label{fig:RCM_compare}
\end{figure*}

The Bayesian high-resolution Transformer not only produces more reliable uncertainty maps, but also can improve model performance. For the SAR datasets, the high-resolution Transformer is capable of detecting small sea ice features like leads and floes. Table \ref{tab:feature_acc} compares the weighted overall accuracy, ice accuracy, and water accuracy for each VI method, as well as the deterministic Transformer model and NASA Team SIC. The high-resolution Transformer-based models all achieved higher feature detection accuracy than the NASA Team SIC. The Bayesian Transformer achieved the highest overall accuracy for Sentinel-1, while the deterministic Transformer outperformed the other methods for RCM. For both SAR datasets, the Bayesian Transformer achieved the highest ice detection accuracy. However, this improvement was accompanied by reduced water accuracy, suggesting a higher sensitivity to ice signatures.

Figures \ref{fig:S1_compare} and \ref{fig:RCM_compare} provide a small-scale visual comparison of the SIC estimates across each method for Sentinel-1 and RCM, respectively. These comparisons further proved that the stochastic model parameters in the proposed Bayesian Transformer could also improve sea ice feature detection. Ice floes approximately 1 or 2 pixels wide have been detected by the Bayesian Transformer, demonstrating resolution up to 200 m for the high-resolution SAR models. The proposed Bayesian Transformer shows consistency with the NASA Team SIC (Figure \ref{fig:S1_compare} row 3 and Figure \ref{fig:RCM_compare} row 3) while also identifying the smallest ice features in the SAR imagery (Figure \ref{fig:S1_compare} row 4 and Figure \ref{fig:RCM_compare} row 2). Additionally, the proposed Bayesian Transformer mitigated the impact of wind effects (Figure \ref{fig:S1_compare} row 3) and thermal noise (Figure \ref{fig:RCM_compare} row 1, row 2, and row 3) on SIC estimates. This shows how regularization through priors and explicit uncertainty estimation during learning with the Bayesian Transformer can effectively reduce overfitting to noisy or inexact training samples for SIC mapping. Overall, all methods that utilize the Transformer model produced higher resolution SIC than the original NASA Team SIC product and preserved many of the small ice features that are present in the SAR imagery, as supported by the quantitative comparison in Table \ref{tab:feature_acc}.

\subsubsection{Model Architecture Comparison}

\begin{table}[]
\renewcommand{\arraystretch}{1.2}
\centering
\caption{Model Architecture Comparison Overall Accuracy/Ice Accuracy/Water Accuracy for Feature Detection. Best performance is \textbf{bolded}.}
\begin{tabular}{l|l|l}
\toprule
Model            & Sentinel-1     & RCM            \\ \hline
U-Net            & 0.69/0.80/0.58 & 0.58/0.71/0.44 \\ \hline
HRNet            & 0.64/0.72/0.56 & 0.61/0.89/0.32 \\ \hline
ConvNeXt         & 0.71/0.77/0.65 & 0.62/0.77/\textbf{0.46} \\ \hline
Swin Transformer & 0.63/0.75/0.52 & 0.60/0.85/0.36 \\ \hline
SegFormer        & 0.64/0.77/0.51 & 0.61/0.79/0.42 \\ \hline
\textbf{HR Transformer}   & \textbf{0.76/0.86/0.65} & \textbf{0.65/0.91/}0.40 \\
\bottomrule
\end{tabular}
\vspace{-0.3cm}
\label{tab:model_comparison}
\end{table}

The high-resolution (HR) Transformer model architecture was evaluated against CNN-based models, i.e., U-Net \cite{ronneberger2015unet}, HRNet \cite{wang2020deep}, and ConvNeXt \cite{liu2022convnet2020s}, and Transformer-based models, i.e., Swin Transformer \cite{Liu_2021_ICCV} and SegFormer \cite{xie2021segformer} for pan-Arctic SIC mapping. A single day was used for training (2021/09/10) and the validation dataset was used to determine the best models. Table \ref{tab:model_comparison} compares Sentinel-1 and RCM-based feature detection accuracy for each model according to the same 300 high quality validation samples used in Section \ref{sec:local_acc}. The highest overall accuracy and ice accuracy were achieved by the high-resolution Transformer for both Sentinel-1 and RCM. The ConvNeXt model achieved the next best results, providing high water detection accuracy for both SAR systems. These feature detection results align with Section \ref{sec:local_acc}, demonstrating that the high-resolution Transformer model performs particularly well in ice identification, which is critical for safe navigation.

\begin{figure*}[!t]
    \centering
    \includegraphics[width=0.9\linewidth]{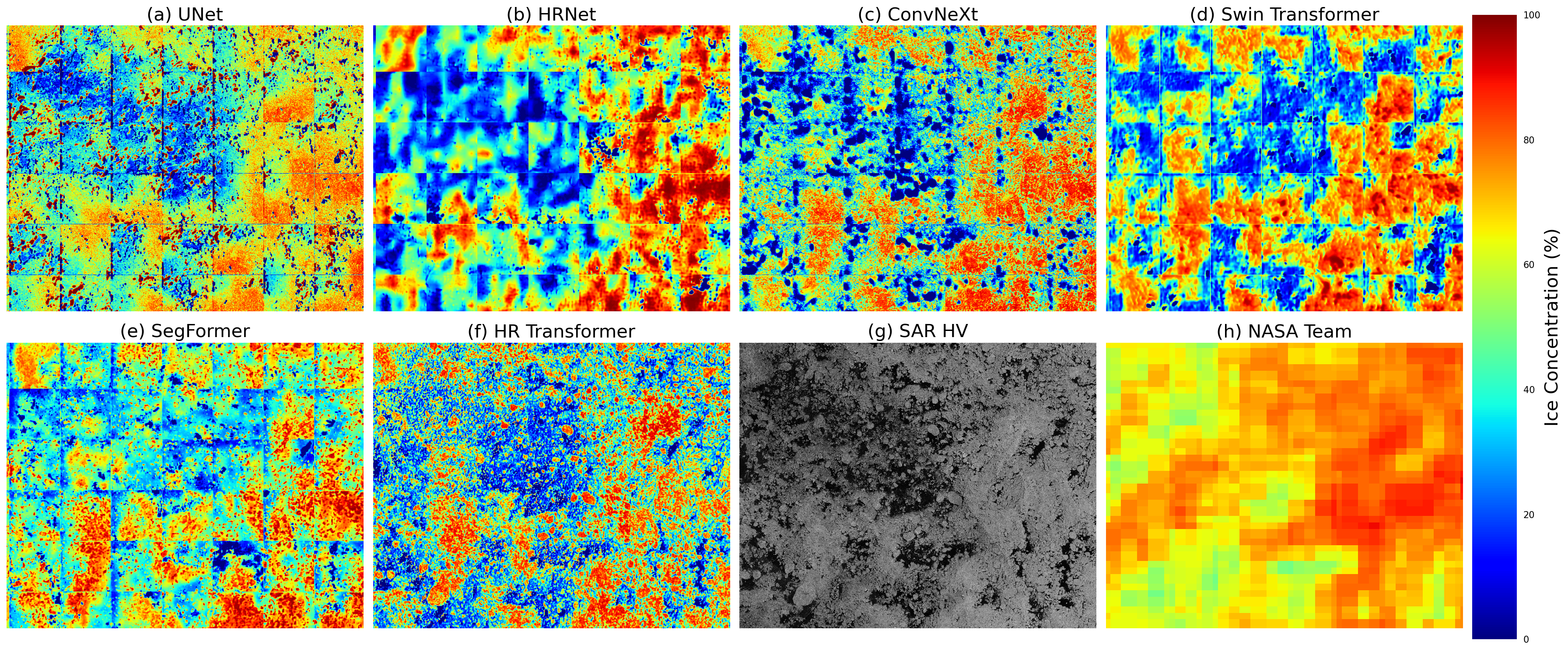}
    \vspace{-0.2cm}
    \caption{Local visual comparison of SIC in \textbf{MIZ} region derived from Sentinel-1 on September 4th, 2021 using (a) U-Net, (b) HRNet, (c) ConvNeXt, (d) Swin Transformer, (e) SegFormer, and \textbf{(f) High-Resolution (HR) Transformer (our approach)}. SIC predictions are compared to (g) SAR HV and (h) NASA Team. Our approach detects small sea ice floes and ice boundaries in the MIZ while maintaining large-scale SIC patterns.}
    \label{fig:S1_model_compare_miz}
    \includegraphics[width=0.9\linewidth]{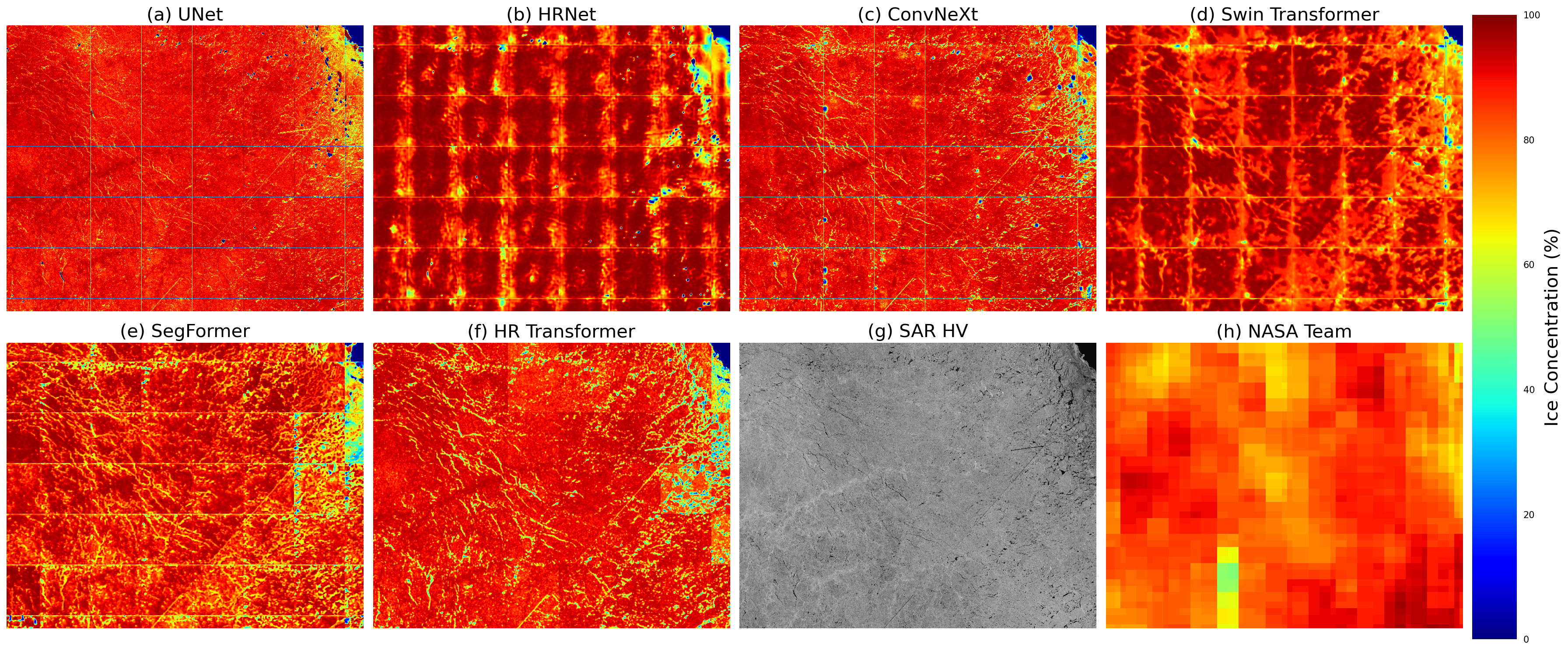}
    \vspace{-0.2cm}
    \caption{Local visual comparison of SIC in \textbf{ice pack} region derived from Sentinel-1 on September 4th, 2021 using (a) U-Net, (b) HRNet, (c) ConvNeXt, (d) Swin Transformer, (e) SegFormer, and \textbf{(f) High-Resolution (HR) Transformer (our approach)}. SIC predictions are compared to (g) SAR HV and (h) NASA Team. Our approach detects small leads/cracks in ice pack while maintaining large-scale SIC patterns.}
    \label{fig:S1_model_compare_ice}
    \includegraphics[width=0.9\linewidth]{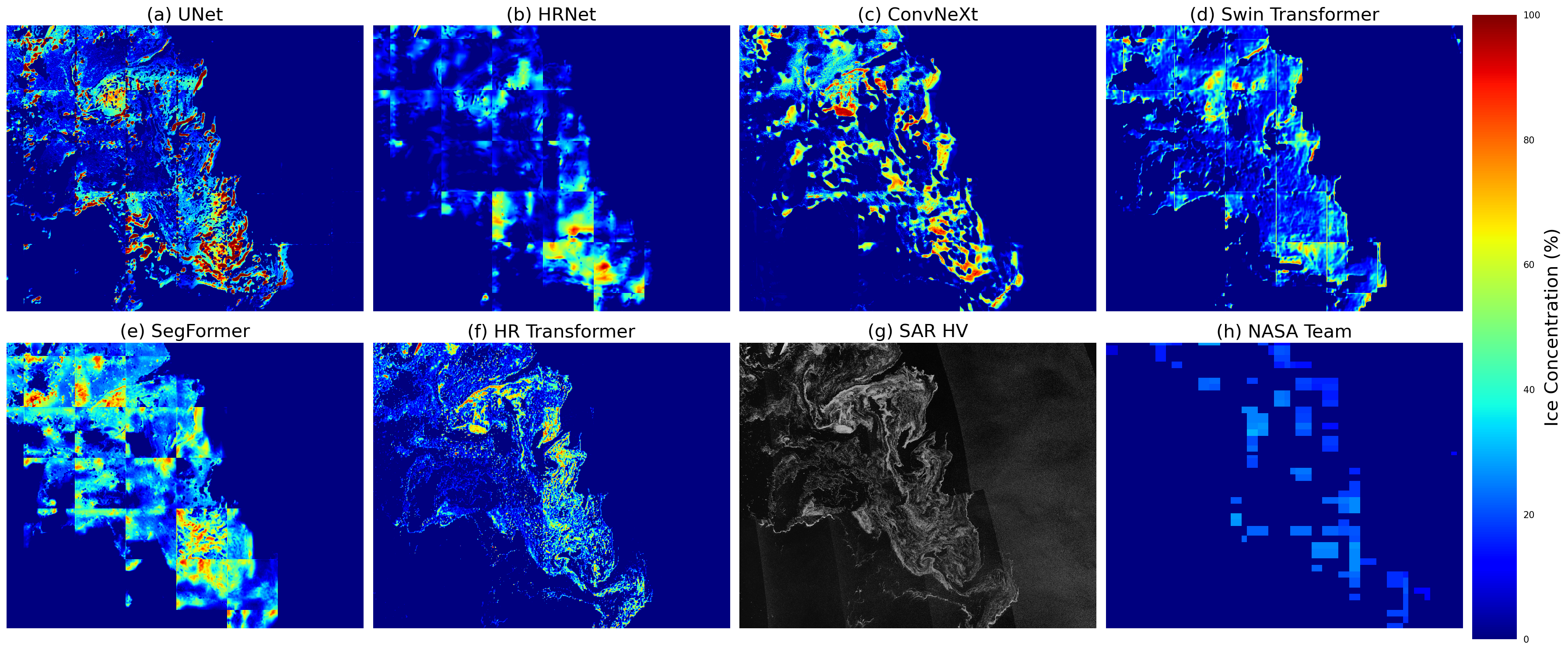}
    \vspace{-0.2cm}
    \caption{Local visual comparison of SIC in \textbf{near open water} region derived from Sentinel-1 on September 4th, 2021 using (a) U-Net, (b) HRNet, (c) ConvNeXt, (d) Swin Transformer, (e) SegFormer, and \textbf{(f) High-Resolution (HR) Transformer (our approach)}. SIC predictions are compared to (g) SAR HV and (h) NASA Team. Our approach detects thin ice without over estimation.}
    \label{fig:S1_model_compare_water}
\end{figure*}

A visual comparison of Sentinel-1-derived SIC in the MIZ, ice pack, and near open water regions for September 4th, 2021 is presented in Figures \ref{fig:S1_model_compare_miz}, \ref{fig:S1_model_compare_ice}, and \ref{fig:S1_model_compare_water}, respectively. Overall, the comparison demonstrates that the high-resolution Transformer model effectively identifies floes and leads while preserving large-scale SIC structure in both heterogeneous MIZ and homogeneous ice pack regions. In contrast, HRNet, Swin Transformer, and SegFormer exhibit spatial inconsistencies in the homogeneous ice pack, as shown in Figure \ref{fig:S1_model_compare_ice}, where block-like artifacts become apparent. This suggests that the high-resolution Transformer more effectively maintains spatial continuity across tiles and mitigates boundary effects, likely due to its dedicated local and global modules.

\begin{figure*}[!t]
    \centering
    \includegraphics[width=0.9\linewidth]{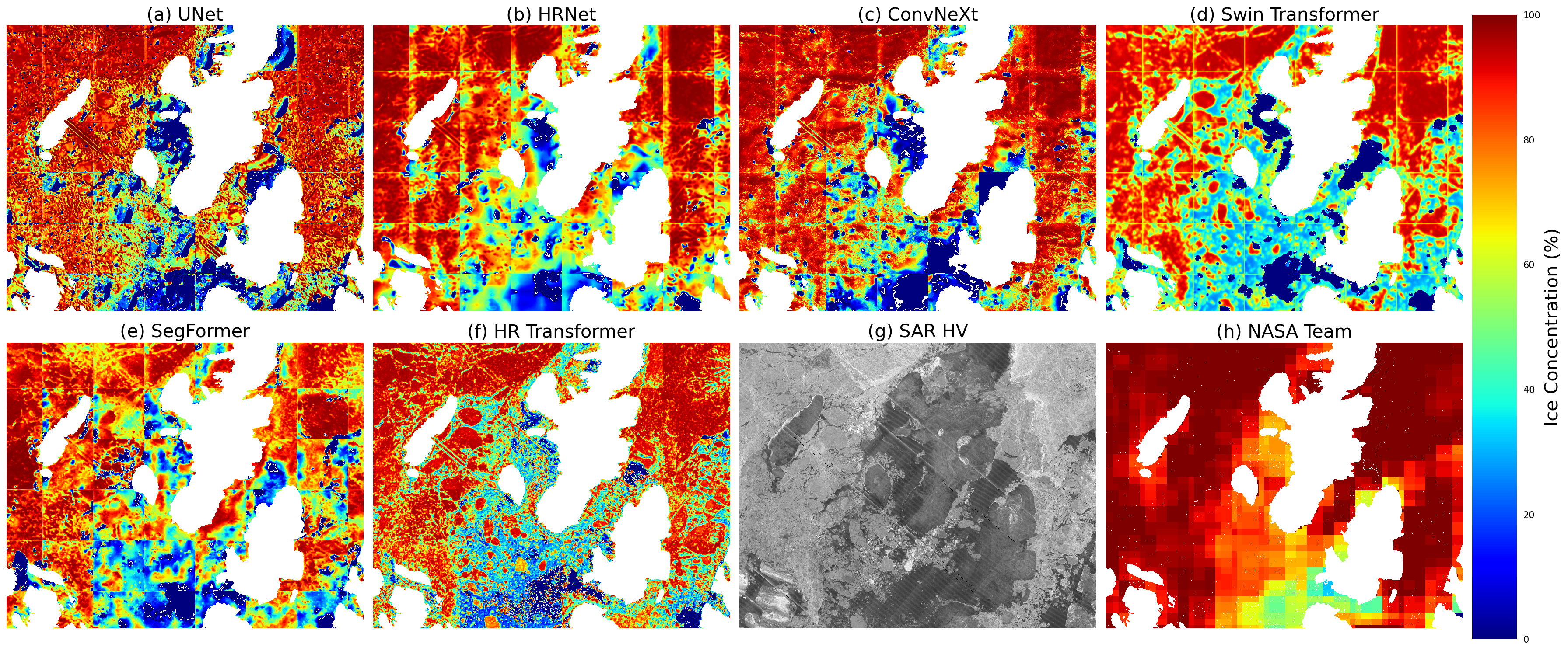}
    \vspace{-0.2cm}
    \caption{Local visual comparison of SIC in \textbf{MIZ} region derived from RCM on September 18th, 2021 using (a) U-Net, (b) HRNet, (c) ConvNeXt, (d) Swin Transformer, (e) SegFormer, and \textbf{(f) High-Resolution (HR) Transformer (our approach)}. SIC predictions are compared to (g) SAR HV and (h) NASA Team. Our approach detects small sea ice floes and ice boundaries in the MIZ while maintaining large-scale SIC patterns.}
    \label{fig:RCM_model_compare_miz}
    \includegraphics[width=0.9\linewidth]{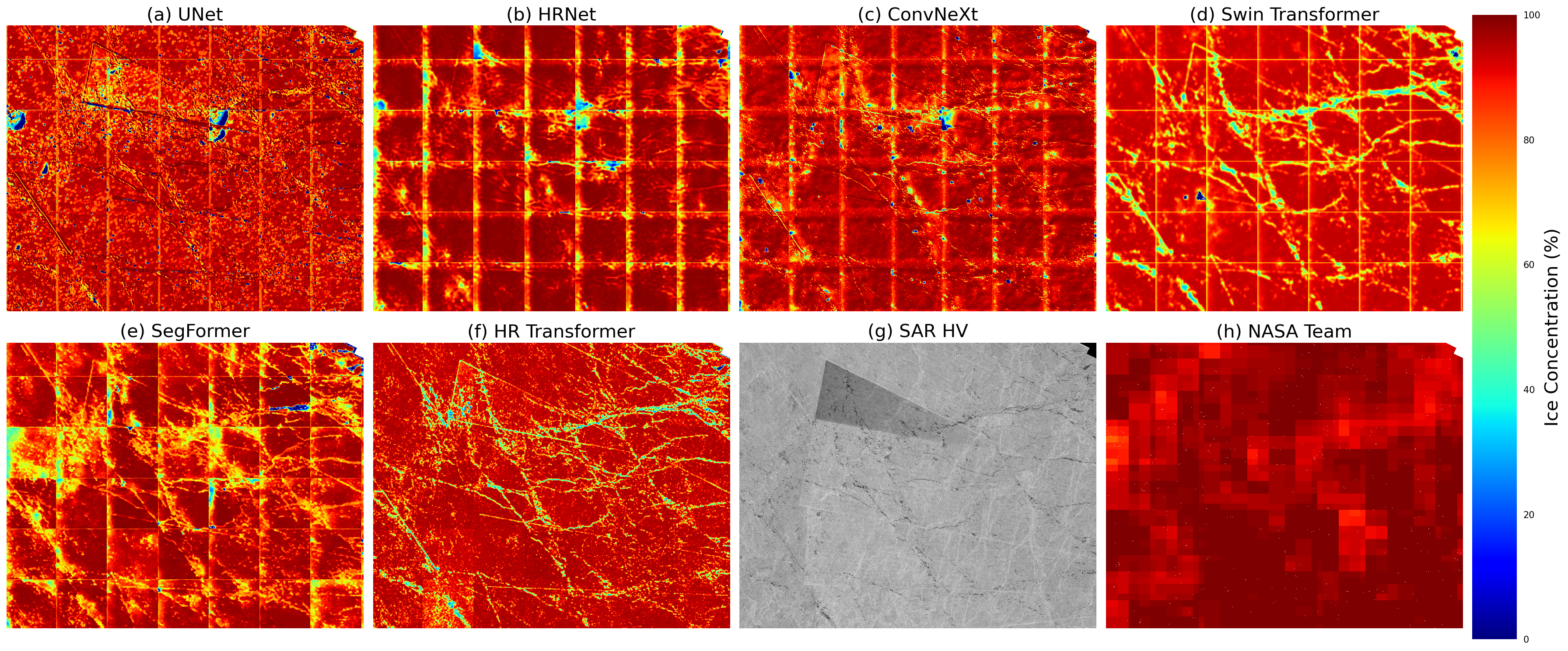}
    \vspace{-0.2cm}
    \caption{Local visual comparison of SIC in \textbf{ice pack} region derived from RCM on September 18th, 2021 using (a) U-Net, (b) HRNet, (c) ConvNeXt, (d) Swin Transformer, (e) SegFormer, and \textbf{(f) High-Resolution (HR) Transformer (our approach)}. SIC predictions are compared to (g) SAR HV and (h) NASA Team.  Our approach detects small leads/cracks in ice pack while maintaining large-scale SIC patterns.}
    \label{fig:RCM_model_compare_ice}
    \includegraphics[width=0.9\linewidth]{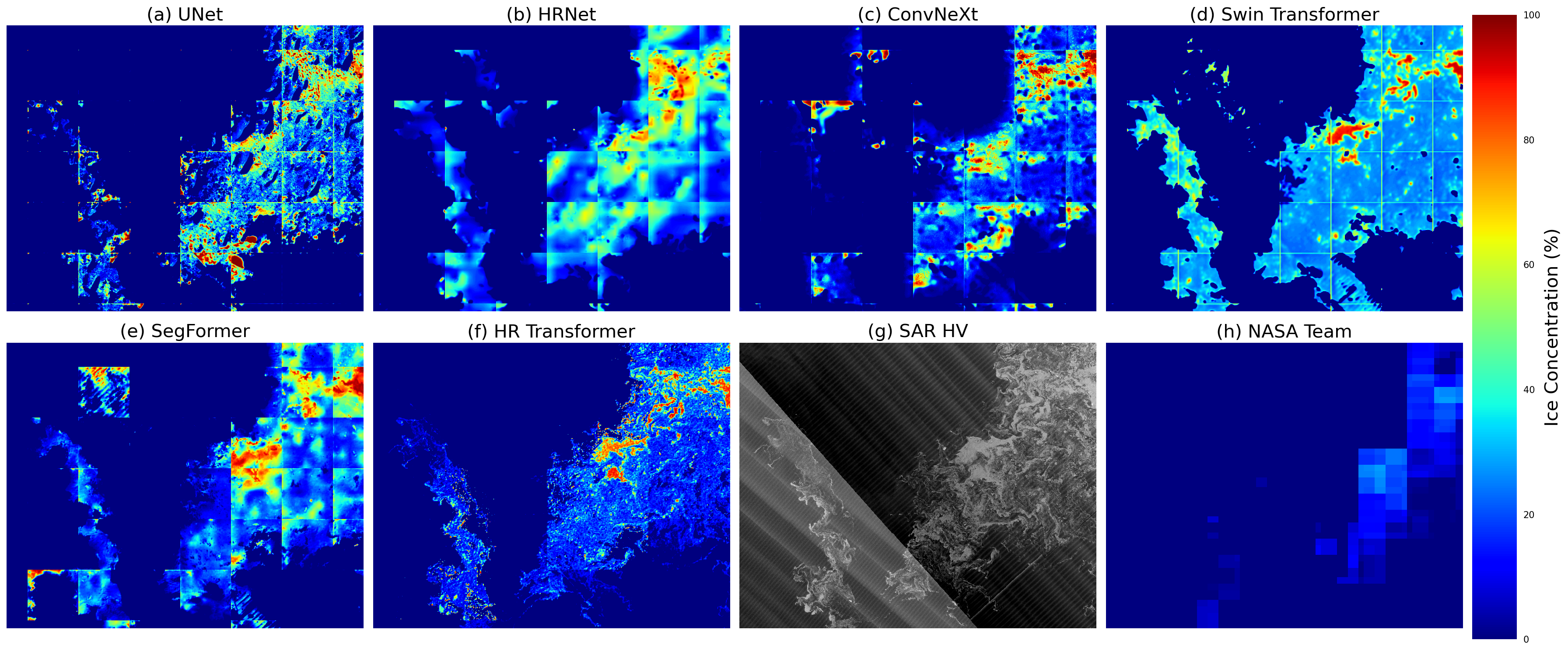}
    \vspace{-0.2cm}
    \caption{Local visual comparison of SIC in \textbf{near open water} region derived from RCM on September 18th, 2021 using (a) U-Net, (b) HRNet, (c) ConvNeXt, (d) Swin Transformer, (e) SegFormer, and \textbf{(f) High-Resolution (HR) Transformer (our approach)}. SIC predictions are compared to (g) SAR HV and (h) NASA Team. Our approach detects thin ice while mitigating the impact of incidence angle effects and thermal banding noise.}
    \label{fig:RCM_model_compare_water}
\end{figure*}

Similar visual model comparison results for RCM-derived SIC on September 18th, 2021 are presented in Figures \ref{fig:RCM_model_compare_miz}, \ref{fig:RCM_model_compare_ice}, and \ref{fig:RCM_model_compare_water}. These examples highlight the additional challenges inherent to the RCM dataset. Figures \ref{fig:RCM_model_compare_ice}(g) and \ref{fig:RCM_model_compare_water}(g) exhibit SAR seam lines associated with conflicting incidence angle effects, while Figure \ref{fig:RCM_model_compare_water}(g) illustrates thermal banding over open water in the HV channel. Despite these artifacts, the high-resolution Transformer maintains spatially consistent SIC patterns and reduces noise in the detection of small-scale sea ice features. In contrast, the SegFormer prediction in Figure \ref{fig:RCM_model_compare_water}(e) shows misclassification associated with thermal banding, whereas the high-resolution Transformer output in Figure \ref{fig:RCM_model_compare_water}(f) demonstrates improved robustness to this effect.

\subsection{Ablation Studies}

\begin{table}[]
\renewcommand{\arraystretch}{1.2}
\centering
\caption{Ablation Study Overall Accuracy/Ice Accuracy/Water Accuracy for Feature Detection. Best performance is \textbf{bolded}.}
\begin{tabular}{l|l|l}
\toprule
Dataset                     & Model                                                                 & Metrics        \\ \hline
\multirow{3}{*}{Sentinel-1} & \begin{tabular}[c]{@{}l@{}}\textbf{HR Transformer}\end{tabular} & \textbf{0.70/0.83/0.57} \\
                            & \begin{tabular}[c]{@{}l@{}}LoFormer Block Removed\end{tabular}    & 0.67/0.80/0.53 \\
                            & \begin{tabular}[c]{@{}l@{}}$\mathcal{L}_{L1-GW}$ Removed\end{tabular}                  & 0.69/0.83/0.54 \\ \hline
\multirow{3}{*}{RCM}        & \begin{tabular}[c]{@{}l@{}}\textbf{HR Transformer}\end{tabular} & \textbf{0.69/0.90/0.48} \\
                            & \begin{tabular}[c]{@{}l@{}}LoFormer Block Removed\end{tabular}    & 0.63/0.87/0.39 \\
                            & \begin{tabular}[c]{@{}l@{}}$\mathcal{L}_{L1-GW}$ Removed\end{tabular}                  & 0.62/0.90/0.34 \\ \hline
\multirow{3}{*}{AMSR2}      & \begin{tabular}[c]{@{}l@{}}\textbf{HR Transformer}\end{tabular} & \textbf{0.56}/0.78/\textbf{0.33} \\
                            & \begin{tabular}[c]{@{}l@{}}LoFormer Block Removed\end{tabular}    & 0.55/\textbf{0.81}/0.29 \\
                            & \begin{tabular}[c]{@{}l@{}}$\mathcal{L}_{L1-GW}$ Removed\end{tabular}                  & 0.55/0.79/0.30\\
\bottomrule
\end{tabular}
\vspace{-0.3cm}
\label{tab:feature_ablation_acc}
\end{table}

Ablation studies were performed to assess key innovations in the proposed high-resolution Transformer architecture and training strategy in Figure \ref{fig:Transformer}(a) and (c), respectively. For each dataset, the impact of an individual component on feature detection performance was evaluated by retraining the high-resolution Transformer with the component removed. Tables \ref{tab:feature_ablation_acc} compares feature detection accuracy for each experiment using the same high quality samples as Section \ref{sec:local_acc}. When comparing the high-resolution Transformer feature detection accuracy across datasets, Sentinel-1 achieved the highest overall accuracy (0.70), followed by RCM (0.69) and AMSR2 (0.56). This comparison supports our choice to layer Sentinel-1 first, followed by RCM and AMSR2, during decision-level fusion.

To isolate its impact on fine-scale feature learning, the models were retrained without the LoFormer block. As shown in Table \ref{tab:feature_ablation_acc}, for the SAR datasets removing the LoFormer block resulted in lower ice and water detection accuracy. This demonstrates the essential role of the LoFormer in capturing high-frequency details in the SAR imagery that the global attention mechanism in the GloFormer block tends to overlook. However, for the AMSR2 dataset, removing the LoFormer block resulted in higher ice detection accuracy. This indicates that for the AMSR2 dataset, the small-scale features in the 89.0 GHz channels may be overpowered by atmospheric interference. Additionally, as the LoFormer block introduces higher model complexity, the model may be overfitting to the low resolution AMSR2 training data. Therefore, the high-frequency focus of the LoFormer is scale-dependent. While critical for high-resolution SAR imagery, it can introduce inductive biases that are counterproductive for the coarse spatial resolution and inherent noise characteristics of the AMSR2 89.0 GHz channels.

To evaluate the proposed training strategy, the models were retrained using a weakly supervised $\mathcal{L}_{L1}$ loss without geographical weighting. This means that every sample was weighted equally across the open water, ice pack, and MIZ. As shown in Table \ref{tab:feature_ablation_acc}, the exclusion of geographical weights led to a decrease in overall sea ice feature detection accuracy, primarily driven by the decline in water detection performance for all three datasets. These results suggest that incorporating geographical weighting effectively mitigated the overestimation of sea ice within the heterogeneous MIZ. As the geographically-weighted $\mathcal{L}_{L1-GW}$ loss placed higher weights to more homogeneous regions, it allows for improved sea ice signature learning from the ice pack and open water while reducing bias caused by noise and signature ambiguity. 

\subsection{Pan-Arctic SIC Accuracy Assessment}

\begin{figure}[]
    \centering
    \includegraphics[width=\linewidth]{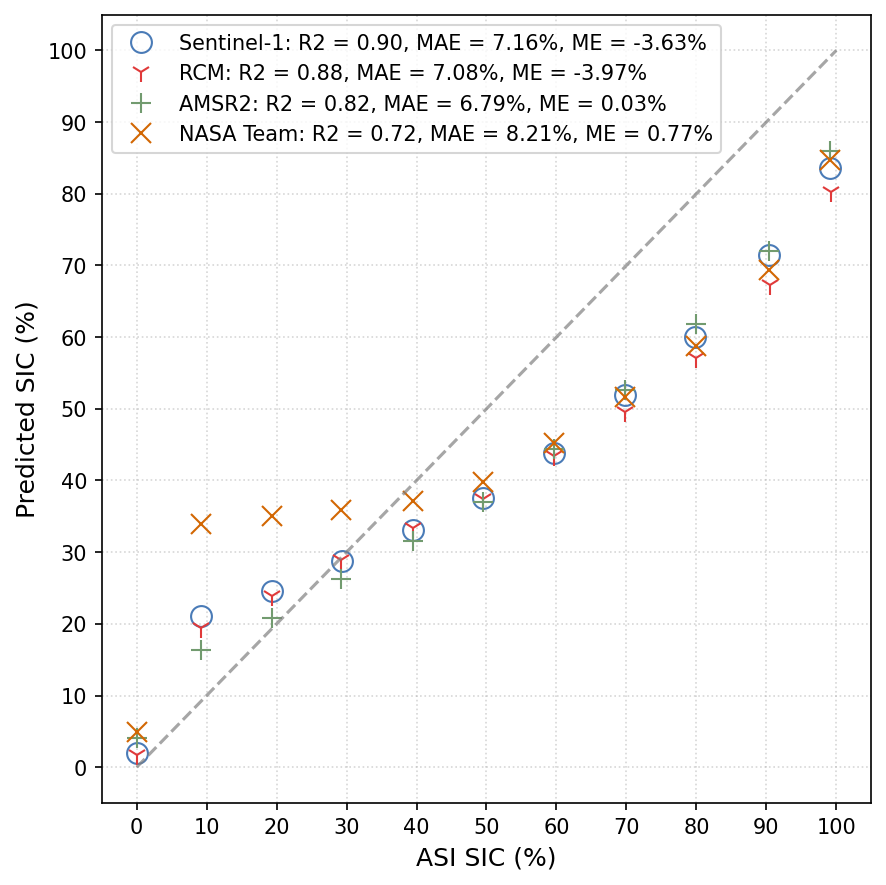}
    \vspace{-0.5cm}
    \caption{Predicted Bayesian Transformer SIC accuracy vs. ASI SIC scatter plot for all \textbf{2021 validation} samples. SIC estimates are aggregated into 11 bins ranging from 0\% to 100\% for visualization, while the R2, RMSE, and MAE are calculated using all individual SIC estimates.}
    \label{fig:asi_validation}
\end{figure}

\begin{figure}[]
    \centering
    \includegraphics[width=\linewidth]{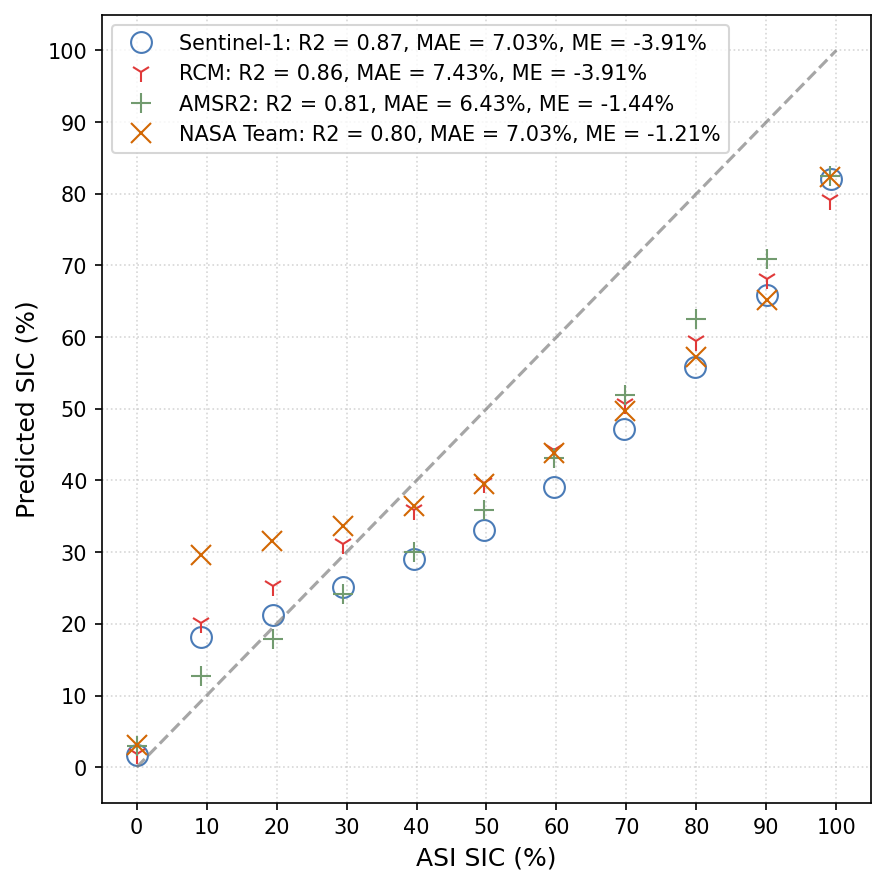}
    \vspace{-0.5cm}
    \caption{Predicted Bayesian Transformer SIC accuracy vs. ASI SIC scatter plot for all \textbf{2025 test} samples. SIC estimates are aggregated into 11 bins ranging from 0\% to 100\% for visualization, while the R2, RMSE, and MAE are calculated using all individual SIC estimates.}
    \label{fig:asi_test}
\end{figure}

\begin{figure}[]
    \centering
    \includegraphics[width=\linewidth]{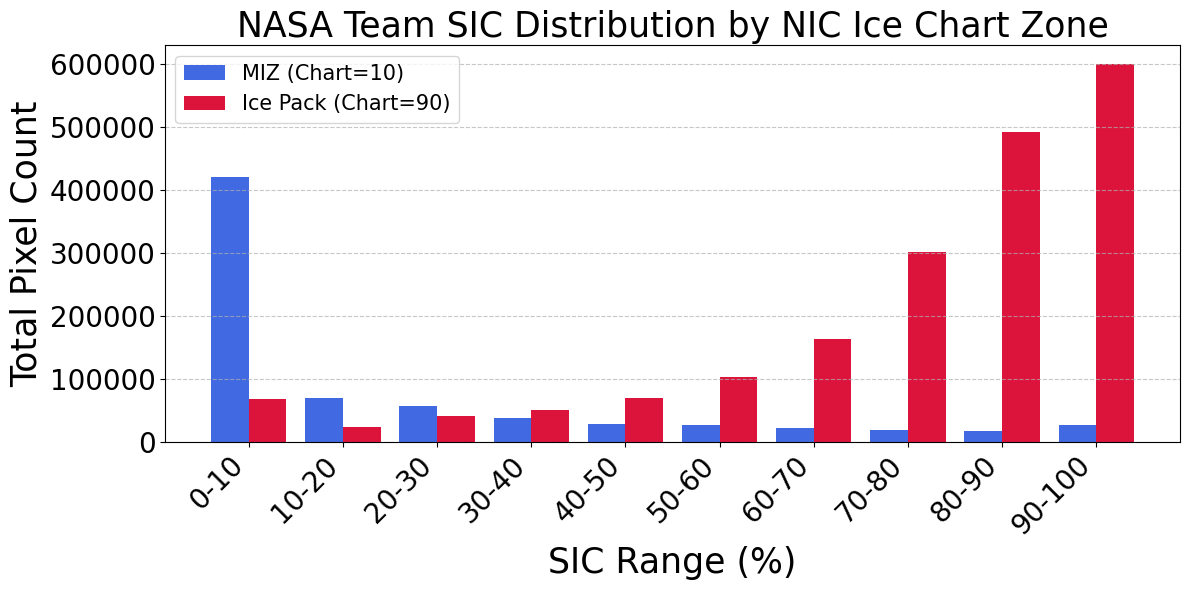}
    \vspace{-0.5cm}
    \caption{Training data distribution of NASA Team SIC in MIZ and Ice Pack according to the NIC Ice Chart.}
    \label{fig:ice_chart_distribution}
\end{figure}

Across data types, Sentinel-1 achieved the highest pan-Arctic SIC accuracy according to R\textsuperscript{2}, followed by RCM and AMSR2 for both validation and test datasets. This measure of overall model performance further supports our choice to layer Sentinel-1 followed by RCM and AMSR2 during decision-level fusion. AMSR2 89.0 GHz SIC predictions achieved the lowest MAE score for both validation and test datasets, demonstrating how resolution and data source consistency can impact accuracy. Both the AMSR2 high-resolution Transformer and ASI algorithm utilize the 89.0 GHZ channels with 3 x 5 km resolution. Generally, for all three data types the R\textsuperscript{2} and MAE demonstrates that the Bayesian high-resolution Transformer outperforms the NASA Team SIC. Assessing the models for systematic bias, the SAR-based models have the most notable bias of approximately 4\% SIC overestimation for both validation and test datasets according to the ME scores. This is most likely caused by the additional high-resolution detail provided by SAR systems.

From 50\% to 100\% ice concentration the model relied on the NASA Team SIC product, resulting in underestimation in this range. However, the Bayesian Transformer outperformed the NASA Team product in the 10\% to 40\% range, indicating that the proposed training strategy is successfully mitigating the impact of NASA Team limitations through geographically-weighted weak supervision. As the MIZ is weighted less during training, it allowed for the model to focus on the strong ice and water signatures in the ice pack and open water, respectively. Figure \ref{fig:ice_chart_distribution} illustrates how the MIZ training samples are dominant within the 0\% to 40\% range, confirming that the reduced weighting corresponds with the regions of improved accuracy. This yielded higher accuracy than NASA Team SIC for both the validation and test datasets. Thus, the geographically-weighted $\mathcal{L}_{L1-GW}$ loss is a valuable tool for addressing sea ice signature ambiguity.

There was stronger agreement across Sentinel-1, RCM, and AMSR2 within the validation datasets compared to the test datasets. However, the overall trends in SIC accuracy remain similar, with the test datasets exhibiting slightly lower overall accuracy. While the proposed decision-level fusion approach can highlight the strengths across different data sources, it does not allow for cross-sensor perception to enforce consistency. However, the parallel branches dedicated to each individual sensor could support development in late stage feature-level fusion with both unimodal and multimodal encoders \cite{bigdeli2021ensemble, jakubik2025terramind}. 

It is also important to note that although the proposed methods demonstrate interannual consistency during the September sea ice minimum, future work is required to assess performance across melt and freeze-up seasons. Many existing approaches instead prioritize cross-seasonal generalization \cite{wang_seasonal_2025, wulf2024panarctic}. Rather than pursuing a single model that performs uniformly across all conditions, a specialized approach such as the proposed Bayesian Transformer could serve as a pre-trained expert within a large-scale mixture-of-experts (MoE) framework \cite{ding2025denseformer, chen2025heterogeneous}. Leveraging a collection of highly specialized models, along with the vast amount of heterogeneous satellite imagery collected daily, may ultimately support the development of more robust pan-Arctic foundation models.

\section{Conclusion}
\label{conclusion}
Accurate daily pan-Arctic SIC mapping with corresponding uncertainty quantification is crucial for monitoring climate change, supporting climate adaptation for Northern communities, and ensuring safe navigation as sea ice extents continue to decrease. However, this is a challenging task due to the subtle nature of ice signature features, inexact SIC labels, model ambiguity, and data heterogeneity. In this study, we proposed a Bayesian High-Resolution Transformer with global (GloFormer) and local (LoFormer) modules for improved feature extraction and a built-in Bayesian framework to better capture uncertainty compared to alternative approaches. An innovative geographically-weighted weakly supervised training strategy was proposed to achieve fine-scale SIC estimates from low resolution and inexact labels. We performed decision-level fusion of Sentinel-1, RCM, and AMSR2 data to address data heterogeneity and enhance pan-Arctic SIC mapping. These contributions are supported by the following key findings:

\begin{enumerate}
    \item Our high-resolution Transformer architecture produced SIC and uncertainty estimates with spatial resolution up to 200 m, enabling the detection of small ice leads and floes while preserving large-scale SIC patterns consistent with the ASI SIC algorithm.
    \item The proposed geographically-weighted weakly supervised $\mathcal{L}_{L1-GW}$ loss enhanced spatial resolution and improved SIC accuracy compared to the NASA Team SIC used as weak labels. This improvement is most evident in the MIZ, where low resolution SIC labels do not accurately represent the transition from pack ice to open water.
    \item The proposed Bayesian framework produced the most reliable uncertainty estimates with the lowest calibration errors among MC dropout and epoch ensemble generation. It also effectively mitigated variance across heterogeneous datasets and demonstrated stronger generalization capabilities than the deterministic Transformer, allowing for more accurate ice feature detection accuracy.
    \item We achieved pan-Arctic SIC and uncertainty estimates with complete daily coverage using decision-level data fusion. This approach successfully leveraged both high-resolution SAR and full coverage PM imagery without compromising their unique strengths.
\end{enumerate}

Despite these novel contributions, there are several limitations to address in future work. While the high-resolution Transformer is well suited for SAR imagery, it could be further optimized for the AMSR2 89.0 GHz channels to enhance detail without overfitting to atmospheric interference. Future work will also explore using more robust operational SIC products, such as the daily ASI products, as weak training labels to enable higher accuracy SIC estimates. To provide more comprehensive uncertainty estimates, both aleatoric (data-driven) and epistemic (model-driven) uncertainties can be modeled through the incorporation of Heteroscedastic Bayesian Neural Networks (HBNN). As the proposed decision-level data fusion approach does not support cross-sensor perception, it could be adapted to late stage feature-level fusion with both unimodal and multimodal encoders. Finally, further model generalization will explore how to build a collection of seasonal expert models for year round SIC mapping.
 
High-resolution SIC mapping with reliable uncertainty estimation will support long-term, continuous mapping of pan-Arctic sea ice, providing a consistent baseline for future environmental studies of the Arctic and cryosphere. Overall, this study offers a first step towards operational high-resolution pan-Arctic sea ice charting.

\section{Acknowledgement}
The authors acknowledge the Government of Canada Earth Observation Data Management System (EODMS), Japan Aerospace Exploration Agency (JAXA), European Space Agency (ESA), and the U.S. National Ice Center for providing the datasets in this study. This work was supported by the Natural
Sciences and Engineering Research Council of Canada (NSERC) under Grant RGPIN-2019-06744.

%\bibliographystyle{modeL1-num-names}
\bibliographystyle{cas-model2-names}

\bibliography{uncertainty_ref}



\end{document}